\documentclass[]{skillnet}
\usepackage{fix-cm}
\usepackage{helvet}           

\usepackage{nicefrac}         
\usepackage{siunitx}         

\usepackage{longtable}
\usepackage{booktabs}

\usepackage{tabularx}         
\usepackage{array}           
\usepackage{makecell}         
\usepackage{threeparttable}

\usepackage{wrapfig}          
\usepackage{float}            
\usepackage[export]{adjustbox}

\usepackage{tikz}             
\usepackage{pgfplots}         
\usepackage{pgf-pie}

\usepackage{listings}        
\usepackage{ragged2e}        
\usepackage{comment}

\usepackage{pifont}           
\usepackage{fontawesome}

\usepackage{enumitem}        
\usepackage{titletoc}         
\usepackage{minitoc}          
\usepackage[toc,page,header]{appendix}

\usepackage{url}              
\usepackage[hang,flushmargin]{footmisc}  
\usepackage{xspace}
\newcommand{\dataname}{{SkillNet-Gym}\xspace}
\newcommand{\notcheckmark}{\textcolor{black}{\bcmark\kern-1.1ex\raisebox{.7ex}{\rotatebox[origin=c]{125}{--}}}\color{black}}
\definecolor{darkgreen}{RGB}{0,160,0}
\pgfplotsset{compat=1.18}
\newcommand{\bcmark}{\color{blue}{\ding{51}}}
\newcommand{\cmark}{\color{darkgreen}{\ding{51}}}
\newcommand{\xmark}{\color{red}{\ding{55}}}
\definecolor{mydarkbrown}{HTML}{d97557}
\definecolor{sky}{HTML}{A7C9E8}  
\definecolor{lightblue}{RGB}{200, 230, 255}  
\definecolor{headerblue}{RGB}{150, 200, 255}

\newcounter{examplebox}

\makeatletter
\newcommand\blfootnote[1]{%
  \begingroup
  \renewcommand\thefootnote{}\footnote{#1}%
  \addtocounter{footnote}{-1}%
  \endgroup
}
\makeatother
\title{%
  \begin{minipage}[c]{0.08\textwidth}
    \includegraphics[height=2.6em]{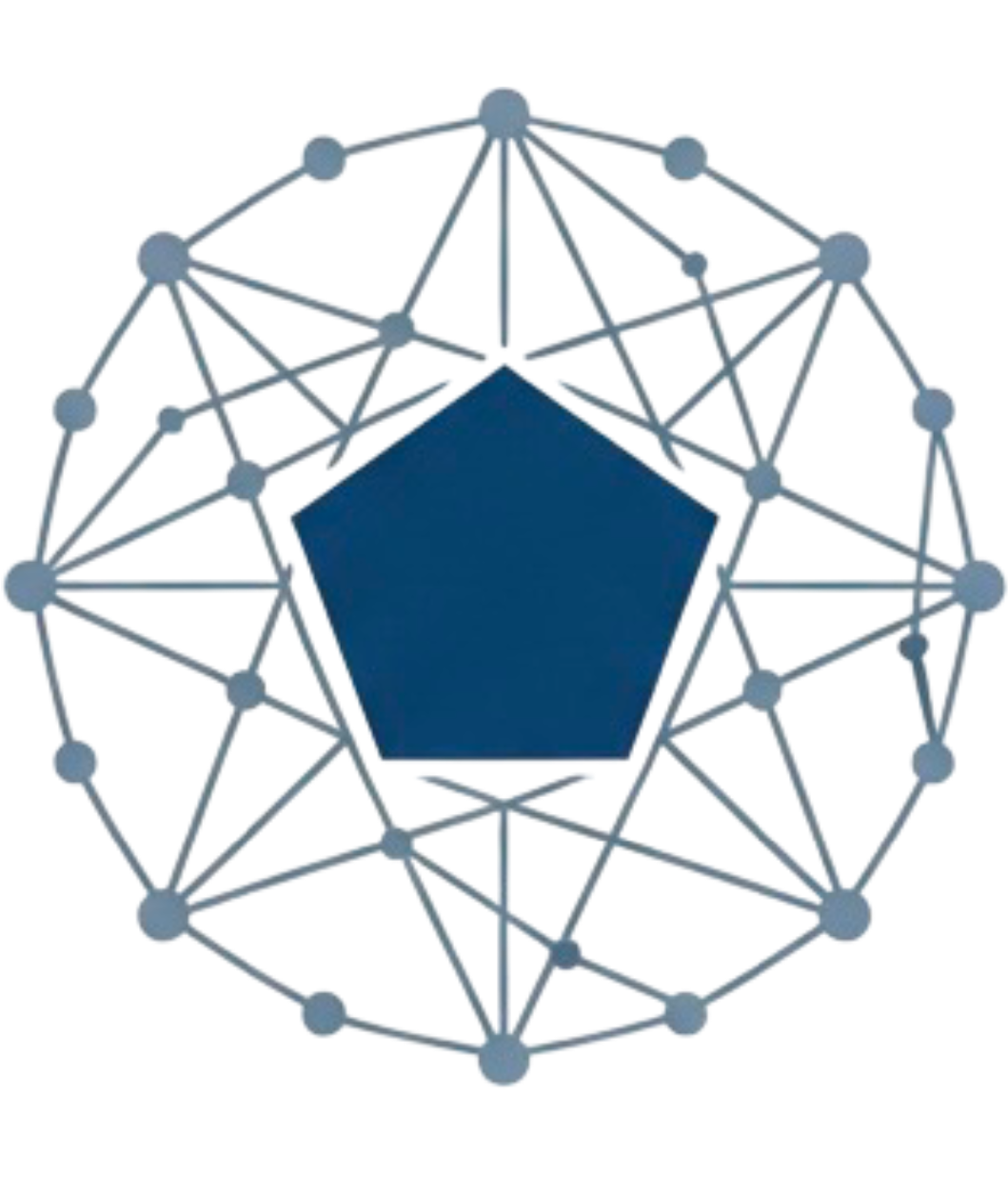} 
  \end{minipage}%
  \hspace{0.8em}%
  \begin{minipage}[c]{0.88\textwidth}
    \centering
    \textsc{SkillNet}: Create, Evaluate, and Connect AI Skills
  \end{minipage}
}

\author{
    Yuan Liang\textsuperscript{1*}, 
    Ruobin Zhong\textsuperscript{1*},   
    Haoming Xu\textsuperscript{1*},
    Chen Jiang\textsuperscript{3*}, 
    Yi Zhong\textsuperscript{1},
     Runnan Fang\textsuperscript{4},
    Jia-Chen Gu\textsuperscript{11},
    Shumin Deng\textsuperscript{15},
    Yunzhi Yao\textsuperscript{1},
    Mengru Wang\textsuperscript{1},
    Shuofei Qiao\textsuperscript{1},
    Yida Xue\textsuperscript{1},
    Xin Xu\textsuperscript{12},
    Tongtong Wu\textsuperscript{14},
    Kun Wang\textsuperscript{16},
    Yang Liu\textsuperscript{16},
    Zhen Bi\textsuperscript{17},
    Jungang Lou\textsuperscript{17},
    Yuchen Eleanor Jiang\textsuperscript{7},
    Hangcheng Zhu\textsuperscript{4},
    Gang Yu\textsuperscript{4},
    Haiwen Hong\textsuperscript{4},
    Longtao Huang\textsuperscript{4},
    Hui Xue\textsuperscript{4},
    Chenxi Wang\textsuperscript{1},
    Yijun Wang\textsuperscript{6},
    Zifei Shan\textsuperscript{6},
    Xi Chen\textsuperscript{6},
    Zhaopeng Tu\textsuperscript{6},
    Feiyu Xiong\textsuperscript{10},\\
    Xin Xie\textsuperscript{5},
    Peng Zhang\textsuperscript{5},
    Zhengke Gui\textsuperscript{5},
    Lei Liang\textsuperscript{5},
    Jun Zhou\textsuperscript{5},
    Chiyu Wu\textsuperscript{8},
    Jin Shang\textsuperscript{8},\\
    Yu Gong\textsuperscript{8},
    Junyu Lin\textsuperscript{9},
    Changliang Xu\textsuperscript{19},
    Hongjie Deng\textsuperscript{1},
    Wen Zhang\textsuperscript{1},
    Keyan Ding\textsuperscript{1},
    Qiang Zhang\textsuperscript{1},
     Fei Huang\textsuperscript{18},
    Ningyu Zhang\textsuperscript{1}$\dagger$,
    Jeff Z. Pan\textsuperscript{13},
    Guilin Qi\textsuperscript{3},
    Haofen Wang\textsuperscript{2},
    Huajun Chen\textsuperscript{1}
}
\affiliation[1]{\mbox{Zhejiang University}}
\affiliation[2]{\mbox{Tongji University}}
\affiliation[3]{\mbox{Southeast University}}
\affiliation[4]{\mbox{Alibaba Group}}
\affiliation[5]{\mbox{Ant Group}}
\affiliation[6]{\mbox{Tencent}}
\affiliation[7]{\mbox{OPPO}}
\affiliation[8]{\mbox{HomologyAI}}
\affiliation[9]{\mbox{Fudan University}}
\affiliation[10]{\mbox{MemTensor (Shanghai) Technology}}
\affiliation[11]{\mbox{UCLA}}
\affiliation[12]{\mbox{UCSD}}
\affiliation[13]{\mbox{The University of Edinburgh}}
\affiliation[14]{\mbox{Monash University}}
\affiliation[15]{\mbox{NUS}}
\affiliation[16]{\mbox{NTU}}
\affiliation[17]{\mbox{Huzhou University}}
\affiliation[18]{\mbox{Honor Device Co., Ltd}}
\affiliation[19]{\mbox{Hangzhou Institute for Advanced Study, UCAS}}

\abstract{

Current AI agents can flexibly invoke tools and execute complex tasks, yet their long-term advancement is hindered by the lack of systematic accumulation and transfer of skills. Without a unified mechanism for skill consolidation, agents frequently ``reinvent the wheel'', rediscovering solutions in isolated contexts without leveraging prior strategies. To address this challenge, we introduce {\bf SkillNet}, an open infrastructure for creating, evaluating, and organizing AI skills at scale. SkillNet structures skills within a unified ontology that supports creating skills from heterogeneous sources, establishing rich relational connections, and performing multi-dimensional evaluation across {\em Safety}, {\em Completeness}, {\em Executability}, {\em Maintainability}, and {\em Cost-awareness}. Our infrastructure integrates a repository of over 600,000 skills, an interactive platform, and a versatile Python toolkit. Experiments on ALFWorld, WebShop, and ScienceWorld show 40\% higher average rewards and 30\% fewer execution steps across multiple backbone models. Furthermore, {\bf SkillNet-Gym} benchmarks skill retrieval, utilization, and composition, while {\bf SkillNet-Fabric} enables task-specific skill routing through lightweight Wikis. By formalizing skills as evolving, composable assets, SkillNet provides a robust foundation for agents to move from transient experience to durable mastery.
}

\correspondence{\email{liang\_yuan@zju.edu.cn}, \email{zhangningyu@zju.edu.cn}}
\checkdata[Website]{\url{http://skillnet.openkg.cn}, \url{https://github.com/zjunlp/SkillNet}}

\begin{document}

\blfootnote{$^*$Equal Contribution.}
\blfootnote{$^\dagger$Corresponding authors.}
\maketitle
\clearpage

% Catalogue (Need \newpage)
% \newpage
% \tableofcontents
% \newpage

\vspace{-1.5em}

\begin{figure}[t]
\centering
\includegraphics[width=0.9\linewidth]{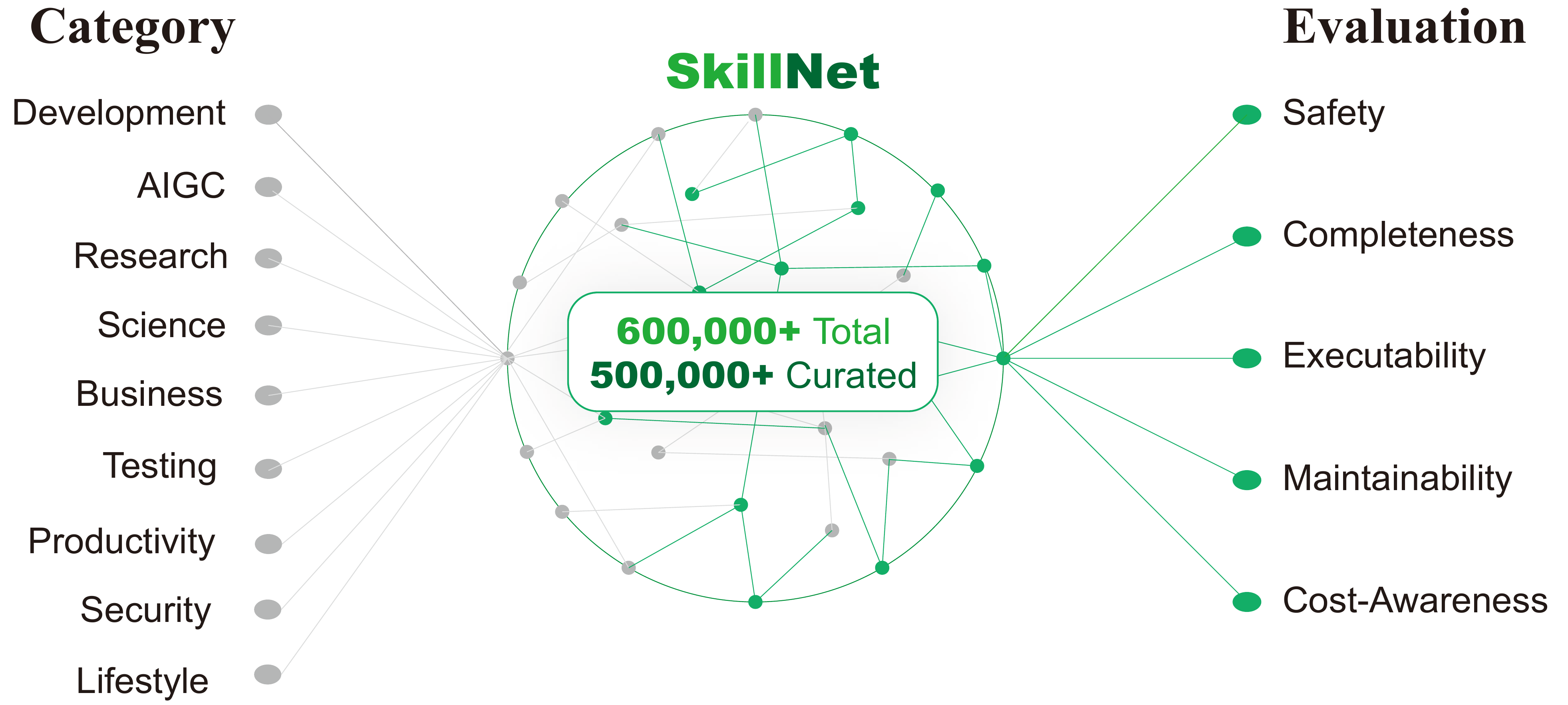}
\caption{Overview of {\bf SkillNet}. SkillNet organizes large-scale agent skills into a structured skill network, modeling rich relations (e.g., similarity, composition, and dependency), supporting multi-dimensional evaluation, and providing unified interfaces for skill discovery, creation, and analysis.}
\label{fig:overview}
\end{figure}
\section{Introduction}

%As Richard S. Sutton noted, ``we are in the era of experience.'' 
%From this perspective, intelligence is grounded not in \emph{ab initio} knowledge acquisition, but in the efficient retrieval and adaptive reuse of heuristics distilled from prior experience \cite{silver2025welcome}. 
%This transition marks the advent of the agent era, where AI moves beyond static question answering to orchestrate long-horizon, executable tasks \cite{chen2022knowprompt,xi2023risepotentiallargelanguage,park2023generativeagentsinteractivesimulacra,wang2023voyageropenendedembodiedagent,zhu2023ghostminecraftgenerallycapable,fang2025synworldvirtualscenariosynthesis,zhao2023survey}.

%Despite recent progress in agentic systems, a critical gap remains. 
%Humans excel at consolidating episodic experience into durable, transferable capabilities, whereas current AI methods largely rely on manual engineering or context learning \cite{dong2024survey,sahoo2024systematic,dou2026cl}. 
%Human cognition abstracts concrete episodes into reusable schemas \cite{hintzman1986schema}: a programmer, for instance, does not merely memorize the syntax of a sorting algorithm but internalizes its underlying logic, enabling transfer across novel codebases without relearning from scratch. 

As Richard S. Sutton noted, ``we are in the era of experience''~\citep{silver2025welcome}. Intelligence is increasingly grounded not in {\em ab initio} knowledge acquisition, but in the efficient retrieval and adaptive reuse of heuristics distilled from prior experience. This transition marks the advent of the agent era, where AI moves beyond static question answering to orchestrate long-horizon, executable tasks~\citep{chen2022knowprompt,xi2023risepotentiallargelanguage,park2023generativeagentsinteractivesimulacra,wang2023voyageropenendedembodiedagent,zhu2023ghostminecraftgenerallycapable,fang2025synworldvirtualscenariosynthesis,zhao2023survey}.
Despite recent progress in agentic systems, a critical yet underexplored question remains: {\bf How can agents systematically consolidate episodic experience into durable, transferable mastery?}

Current AI methods largely rely on manual engineering or transient in-context learning~\citep{dong2024survey,sahoo2024systematic,dou2026cl}. 
While humans excel at consolidating episodic experience into reusable schemas~\citep{hintzman1986schema}, for instance when a programmer internalizes the logic of an algorithm rather than merely memorizing its syntax, contemporary AI systems struggle to bridge the divide between transient context and long-term capability. 
Without a unified mechanism for skill consolidation and sharing, agents repeatedly ``reinvent the wheel'' in isolated contexts, and even well-established strategies rarely inform future tasks.
This gap highlights a broader, longstanding tension in AI between explicit structure and scalable representation, a limitation that becomes especially evident when examined through the historical evolution of knowledge engineering \cite{studer1998knowledge}:
\begin{itemize}
    \item \textbf{The Symbolic Era}: Systems relied on rigid symbolic logic, offering interpretability but suffering from brittleness and limited scalability \cite{lewis1959symbolic,baldoni2018survey,hogan2021knowledge}.
    \item \textbf{The Deep Learning Era}: Knowledge became parametric, taking the form of high-dimensional weight matrices. 
    While powerful, such representations are opaque and difficult to modularize or reuse \cite{lecun2015deep,he2016deep}.
    \item \textbf{The Agentic Era (Current Frontier)}: We are witnessing a convergence toward agent skills, which serve as simple, transferable units that provide agents with new capabilities and expertise while separating intelligence from monolithic parameter spaces \cite{shinn2023reflexionlanguageagentsverbal,zhao2024expelllmagentsexperiential,zheng2025skillweaver,li2026single}.
\end{itemize}

This historical trajectory suggests a natural progression toward skills as modular, externalized units of knowledge that reconcile structural interpretability with scalable representation.
Consequently, the central challenge in the agent era is no longer merely learning from experience, but transforming fragmented experience into durable, composable skill units that support generalizable intelligence. 
However, current AI approaches fall short in two fundamental respects.
First, there is no unified mechanism to acquire and consolidate skills from experience. Valuable expertise exists at scale across open-source repositories, academic publications, and agent execution traces \cite{ye2026metacontextengineeringagentic,chen2026cuaskilldevelopskillscomputer}, yet it remains largely unstructured and disconnected. Unlike human learners, who continuously internalize external information into organized knowledge schemas, AI agents cannot automatically distill these resources into reusable, executable capabilities. As a result, skill acquisition remains a manual and episodic process rather than an autonomous, cumulative one \cite{xi2023risepotentiallargelanguage}.
Second, there is no principled framework to validate and maintain skill quality at scale. Without intrinsic and systematic evaluation, skill repositories are prone to ``pollution,'' where executability, safety, and robustness are assessed only indirectly through downstream task success \cite{skillsmp2026,skillhub2026,qin2023toolllmfacilitatinglargelanguage}. Such stochastic and opaque validation introduces technical debt, undermining long-term capability growth and preventing agents from leveraging accumulated experience in a reliable and scalable manner.
Together, these gaps highlight the need for a structured framework that not only captures and organizes skills, but also rigorously evaluates and interconnects them to enable dependable, large-scale knowledge reuse.

To address these issues, we introduce {\bf SkillNet} (Figure~\ref{fig:overview}), an open infrastructure for creating, evaluating, and organizing AI skills at scale. 
We conceptualize a skill as a unified knowledge representation that bridges unstructured language understanding with structured, machine-executable logic. SkillNet constructs a comprehensive Skill Ontology with three interconnected layers: a taxonomic layer for functional categorization, a relational layer encoding dependencies and composition, and a skill-package layer for modular deployment.
To ensure the reliability of these skills, we propose a multi-dimensional evaluation framework that assesses skills based on {\bf Safety}, {\bf Completeness}, {\bf Executability}, {\bf Maintainability}, and {\bf Cost-awareness}.

Our approach transforms fragmented experience from diverse sources, including execution trajectories, open-source repositories, and documentation, into a structured network of over 500,000 curated skills.
Experimental results across three text-based simulated environments (ALFWorld, WebShop, and ScienceWorld) demonstrate that agents augmented with SkillNet achieve substantial gains. For instance, our approach improves the average reward by 40\% while reducing interaction steps by 30\% across various backbone models (e.g., DeepSeek V3, Gemini 2.5 Pro, and o4 Mini), validating that systematic skill accumulation effectively enhances agent competence cumulatively rather than episodically.

% 增加skillnet-gym

Our contributions are summarized as follows: 

\begin{enumerate}[leftmargin=10pt]
    \item We introduce SkillNet, a unified framework that transforms fragmented agent experience into a structured network of modular, composable skills with rich relational modeling, serving as a scalable foundation for actionable knowledge engineering.
    \item We establish a rigorous skill evaluation protocol that quantitatively measures safety, completeness, executability, maintainability, and cost-awareness, ensuring the reliability of large-scale skill repositories.
    \item We release an open-source ecosystem, including a repository of over 500,000 curated skills, a Python toolkit, and comprehensive benchmarks, which empirically demonstrate significant performance improvements in agent planning and execution tasks.
    \item We introduce SkillNet-Gym, a dynamic benchmark automatically constructed from real-world skill ecosystems to evaluate skill construction, retrieval, and composition. Experiments across diverse frontier models and agent harnesses reveal that current agents remain limited in open-world skill utilization, particularly in skill retrieval and multi-skill orchestration.
    \item We develop SkillNet-Fabric, a task time routing layer that places a task specific Wiki between the global skill ecosystem and final set formation. Evaluations on routing and downstream benchmarks demonstrate improved routing quality and task performance across the tested settings.
\end{enumerate}

% 为什么  skill 是下一代知识表示  

% 为什么要net   隐喻

\section{Agent Skills}

In the context of agentic systems, \textbf{skills} represent a lightweight, modular, and reusable abstraction for extending the capabilities of AI agents \cite{agentskills2025}. Conceptually, a skill encapsulates procedural knowledge, task-specific instructions, and supporting resources, enabling agents to perform complex tasks more accurately, efficiently, and consistently.

Functionally, skills are organized as structured folders containing a central \texttt{SKILL.md} file, which defines the skill's metadata and detailed instructions. The metadata typically includes the skill's name, a brief description of its purpose, and usage conditions, while the instructions provide step-by-step guidance for execution. Skills may optionally include scripts, templates, documentation, and other resources required to perform tasks, forming a self-contained capability package.

The core purpose of skills is to provide agents with access to reusable procedural knowledge and context-specific information on demand. For example, an agent can leverage a skill to automate a data analysis pipeline, perform domain-specific reasoning, or generate structured outputs such as reports or presentations. By encapsulating these capabilities, skills enable consistent and repeatable workflows, reduce reliance on hard-coded rules or ad hoc prompts, and allow knowledge to be shared across agents, teams, or applications.

Skills operate through a progressive three-step process:
\begin{enumerate}
    \item \textbf{Discovery:} Agents initially load only minimal metadata (e.g., name and description) to identify potentially relevant skills for a given task.
    \item \textbf{Activation:} When a task matches a skill’s description, the agent reads the full instructions from \texttt{SKILL.md} and prepares any associated resources.
    \item \textbf{Execution:} The agent follows the instructions and optionally executes bundled code or utilizes referenced assets to complete the task.
\end{enumerate}

One of the key benefits of skills is that they are self-documenting. 
The embedded instructions and structured metadata make it straightforward to understand, audit, and iteratively improve each capability. 
Fundamentally, a skill serves as \textbf{a unified knowledge representation that integrates entities, relationships, workflows, and executable code, encompassing both textual semantics and symbolic outcomes}. 
This hybrid representation enables skills to bridge unstructured language understanding with structured, machine-executable logic.
Skills are highly extensible, ranging from lightweight textual guidance to complex bundles that include executable programs and supporting assets. 
Their file-based, version-controlled nature ensures portability, allowing the same skill to be shared, reused, and deployed across different agents or systems with minimal adaptation.

Note that by encapsulating domain-specific knowledge and operational workflows in a structured and reusable form, skills provide an effective mechanism for institutionalizing business logic. 
This structured accumulation of expertise not only supports the systematic propagation of domain knowledge, but also enhances agent performance through reusable, reliable, and composable capabilities.
However, in practice, these skills are often fragmented, scattered across scripts, prompts, or isolated workflows, and their quality, reliability, and reusability are rarely evaluated in a systematic way. 
Without a unified framework to organize, verify, and evolve skills, agents cannot fully realize the potential of this modular knowledge, limiting their ability to evolve continuously and adapt to new tasks. 
This motivates the development of SkillNet.
\section{SkillNet}

\subsection{Overview}

%图2   流程图
\begin{figure}[ht] \centering \includegraphics[width=1\textwidth]{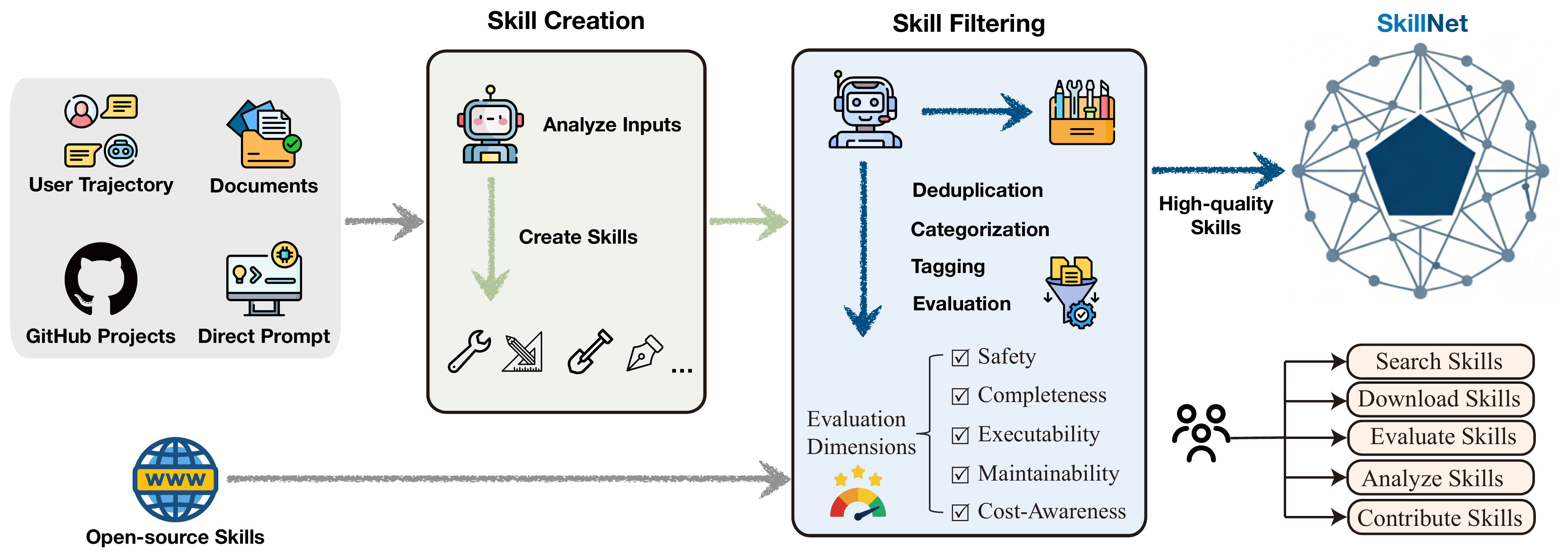} \caption{End-to-end Pipeline of SkillNet. SkillNet transforms heterogeneous user inputs and open internet resources into executable skills through automated skill creation and multi-dimensional evaluation, and organizes high-quality skills into a structured network to support search, download, analysis, and contribution.} \label{fig:fig2} \end{figure}

Figure~\ref{fig:fig2} illustrates the overall architecture of SkillNet, which systematically creates, evaluates, and organizes high-quality skills for agent systems. SkillNet is designed to transform fragmented agent experiences and human knowledge into reusable and verifiable skill entities, enabling scalable and reliable capability growth. SkillNet consists of three core modules:
\begin{figure}[ht] 
    \centering 
    \includegraphics[width=0.9\textwidth]{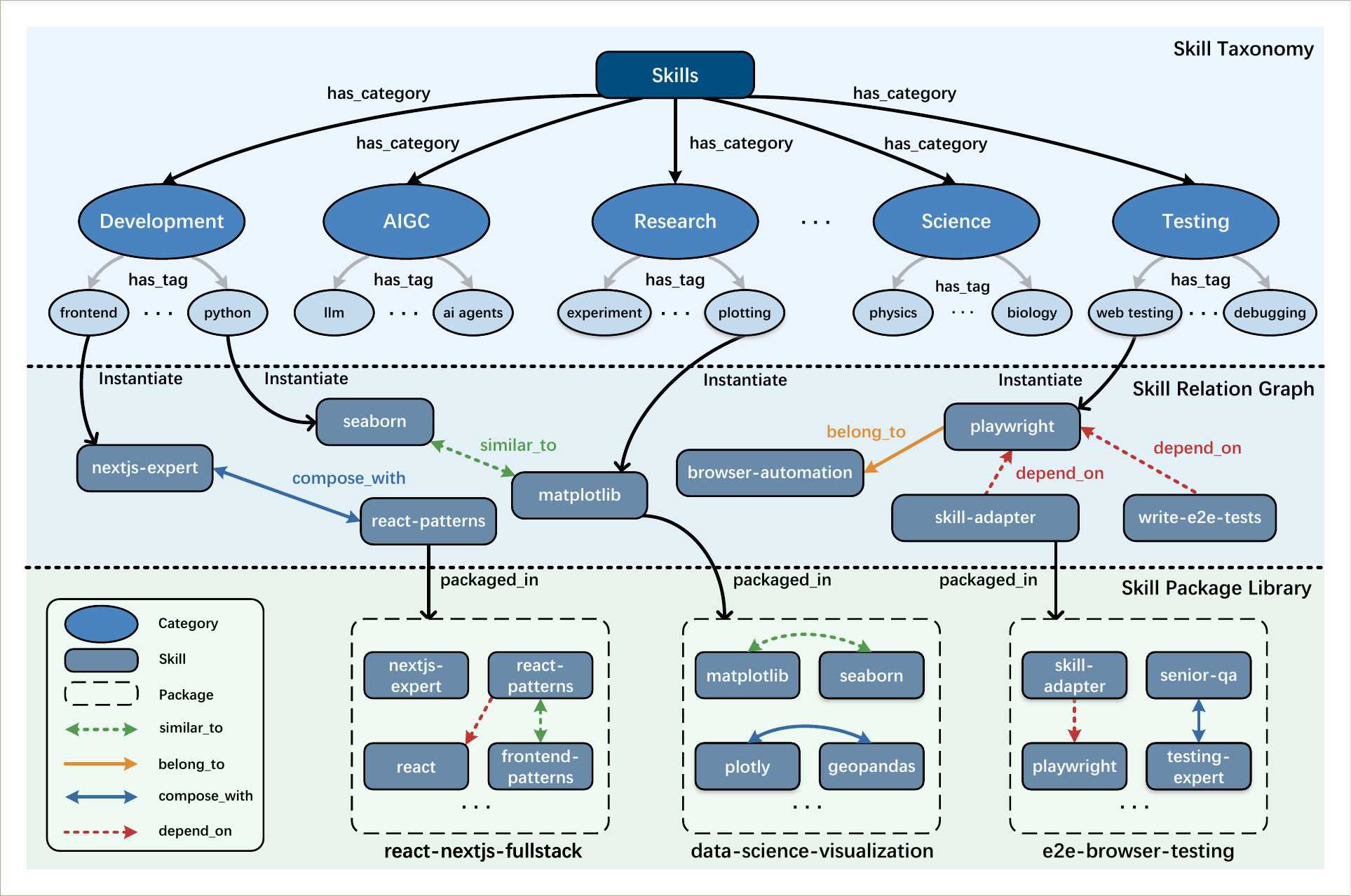} 
    \caption{The Skill Ontology for SkillNet. It consists of three levels: the \textbf{Skill Taxonomy} (top) defines functional categories; the \textbf{Skill Relation Graph} (middle) models inter-skill dependencies and semantic associations; and the \textbf{Skill Package Library} (bottom) organizes skills into modular, task-oriented bundles.} 
    \label{fig:ontology} 
\end{figure}

\paragraph{Skill Creation.} 
This module analyzes diverse inputs, including user trajectories, office documents, GitHub projects, direct prompts, and open internet resources. Based on these inputs, SkillNet generates new skills by extracting executable patterns and structuring them as reusable capabilities.

\paragraph{Skill Evaluation.} 
Generated skills are filtered and evaluated along multiple principal dimensions, including safety, completeness, executability, maintainability, and cost-awareness. This evaluation ensures that only high-quality skills are retained, mitigating redundancy, brittleness, and potential risks.

\paragraph{Skill Analysis.}
Beyond isolated skills, SkillNet automatically analyzes structural and functional relationships among skills, constructing large-scale skill graphs that capture similarity, hierarchy, composition, and dependency patterns. This structured representation enables global reasoning over skill repositories, supporting efficient retrieval, composition, and workflow synthesis.

\paragraph{Open Resources.} 
SkillNet organizes curated, high-quality skills within a structured repository, offering versatile toolkits that enable users and agents to efficiently search, download, create, evaluate, analyze, and contribute skills. 
By providing these standardized interfaces for skill interaction, SkillNet facilitates seamless skill reuse and collaborative evolution across tasks, domains, and agent populations.

\subsection{Skill Ontology}

% The Skill Ontology provides a formal representation of skills, as illustrated in Figure~\ref{fig:ontology}. This ontology definites not only the attributes of individual Skills but also their hierarchical positions and inter-entity relationships within the entire ecosystem. The architecture is organized into three progressive layers:

The Skill Ontology organizes individual skills into a structured, composable network, as illustrated in Figure~\ref{fig:ontology}.
The architecture is organized into three progressive layers:

\paragraph{Skill Taxonomy.} 
This layer organizes skills into a multi-level hierarchy using \texttt{category} and \texttt{tag} relations. It categorizes broad domains (e.g., Development, AIGC, Science ...) into fine-grained tags (e.g., frontend, llm, physics), providing a high-level semantic skeleton.

\paragraph{Skill Relation Graph.} 
This layer instantiates abstract tags into specific \textit{Skill Entities} (e.g., Matplotlib, Playwright). It defines the core interaction logic through multi-relational edges: \texttt{similar\_to}, \texttt{compose\_with}, \texttt{belong\_to}, and \texttt{depend\_on}, forming the backbone for reasoning and planning.

\paragraph{Skill Package Library.} 
The bottom layer represents the physical organization of skills. Individual skills are encapsulated into \textit{Skill Packages} (e.g., \texttt{data-science-visualization}) via the \texttt{packaged\_in} relation, facilitating modular release and deployment.

Note that the skill ontology is dynamic, modeling relations between skills. 
Tags can be continually added from the taxonomy, and LLMs infer relations from them, enabling the instantiation of a skill relation graph.

\subsection{Skill Creation}

SkillNet conceptualizes skills as intermediate capability units that bridge abstract, declarative knowledge and concrete, executable programs. 
Specifically, we develop an automated creation pipeline that systematically transforms heterogeneous information sources into standardized, reusable agent skills.

\subsubsection{From Human Experience to Skill Abstraction}
To construct a comprehensive and versatile skill repository, SkillNet abstracts skills from diverse data sources, transforming heterogeneous information into reusable structured agent skills.

We design a multi-source automated skill creation pipeline that enables SkillNet to induce agentic skills from heterogeneous human knowledge and experience. 
Specifically, SkillNet supports four major categories of data sources: (1) execution trajectories and conversational interaction logs, (2) open-source GitHub repositories, (3) semi-structured documents including PDF, PowerPoint, and Word files, and (4) direct natural language prompts provided by users.
This process is fully implemented through LLMs, and users can also customize the underlying models.
In addition, SkillNet continuously expands its skill pool through open internet resources, in-house development, and community contributions, ensuring sustained scalability and coverage expansion.

\subsubsection{Data-Driven Skill Filtering and Consolidation}
Automatic construction of skills does not imply indiscriminate accumulation. 
Instead, SkillNet introduces a data-driven filtering and consolidation pipeline to ensure the quality of the skill repository. 
Specifically, SkillNet applies a multi-stage curation process consisting of deduplication, filtering, categorization and tagging, evaluation, and final selective consolidation.

Deduplication is performed by jointly comparing skill directory structures and MD5 hashes of skill markdown files, effectively removing redundant skills. Filtering eliminates low-quality, incomplete, or semantically meaningless skills through rule-based validation and model-based checking.
Categorization and tagging classify each skill into one of ten functional categories (Development, AIGC, Research, Science, Business, Testing, Productivity, Security, Lifestyle, Other) and assign fine-grained semantic tags to facilitate retrieval and composition. 
Then, a multi-dimensional evaluation mechanism is applied to determine whether a skill is admitted into the SkillNet repository.
SkillNet automatically establishes inter-skill associations based on ontology-defined relations, ultimately resulting in a structured skill package library.

This structured pipeline enables SkillNet to function as a self-evolving skill ecosystem rather than a static skill collection, continuously improving both skill quality and coverage.

\subsection{Skill Evaluation}

%3. 可验证：技能好不好，如何评估？（Verifiability）

%我们提出一套面向技能的系统性评估指标体系，这是 SkillNet 与现有资源工作的核心差异之一。

%安全性（Safety）
%* 是否存在危险操作？
%* 是否可能被误用或滥用？
%完备性（Completeness）
%* 是否覆盖目标任务的关键步骤？
%* 是否存在隐含前提或缺失条件？
%可执行性（Executability）
%* 给定真实环境与工具，是否能被 Agent 实际执行？
%* 是否存在模糊指令或不可操作步骤？
%可维护性（Maintainability）
%* 是否支持局部修改？
%* 是否可与其他技能安全组合？
%* 版本更新是否会破坏依赖关系？

%成本感知（Cost-awareness）
%这是很多工作没想到，但 reviewer 很爱的一点
%核心问题
%* 执行代价（时间 / 资源 / 金钱）
%在现实 Agent / Tool-using setting 非常现实。

Although skill creation is important, the utility of a skill repository ultimately depends on its reliability. 
To bridge the gap left by existing repositories that lack standardized assessment, we propose a multi-dimensional systematic evaluation framework.
% All evaluation metrics are automatically assessed by the large language model. 
% To validate the reliability of these scores, we recruited three PhD-level annotators to independently compare the model’s ratings with human judgments. 
% The results showed a high degree of agreement between the two, indicating that the model-based evaluation is largely consistent with human assessment.

% We evaluate skills along five distinct dimensions. Safety measures potential risks, such as dangerous system operations (e.g., file deletion) and robustness against misuse or prompt injection attacks. Completeness assesses whether a skill covers all critical steps for task completion and explicitly defines necessary prerequisites and dependencies. Executability goes beyond theoretical correctness by testing whether a skill can be successfully executed using a sandbox environment, including checks for ambiguous instructions or hallucinated tool calls. Maintainability evaluates the skill's modularity to support long-term evolution, examining whether it can be locally modified without breaking global dependencies, whether it is compatible with composition with other skills, and whether updates maintain backward compatibility. Finally, Cost-awareness addresses practical deployment constraints by quantifying execution "price" in terms of latency, computational resources, and API costs, which is essential for optimizing agent efficiency in resource-constrained settings.

We define five core dimensions to quantitatively characterize the quality and readiness of each skill:
\begin{itemize}
\item \textbf{Safety:} Assesses potential risks, including hazardous system operations (e.g., unauthorized file deletion) and robustness against prompt injection or adversarial manipulation.

\item \textbf{Completeness:} Evaluates whether the skill encapsulates all critical procedural steps and explicitly defines necessary prerequisites, dependencies, and execution constraints.
\begin{figure}[htbp]
  \centering
  \includegraphics[width=\textwidth]{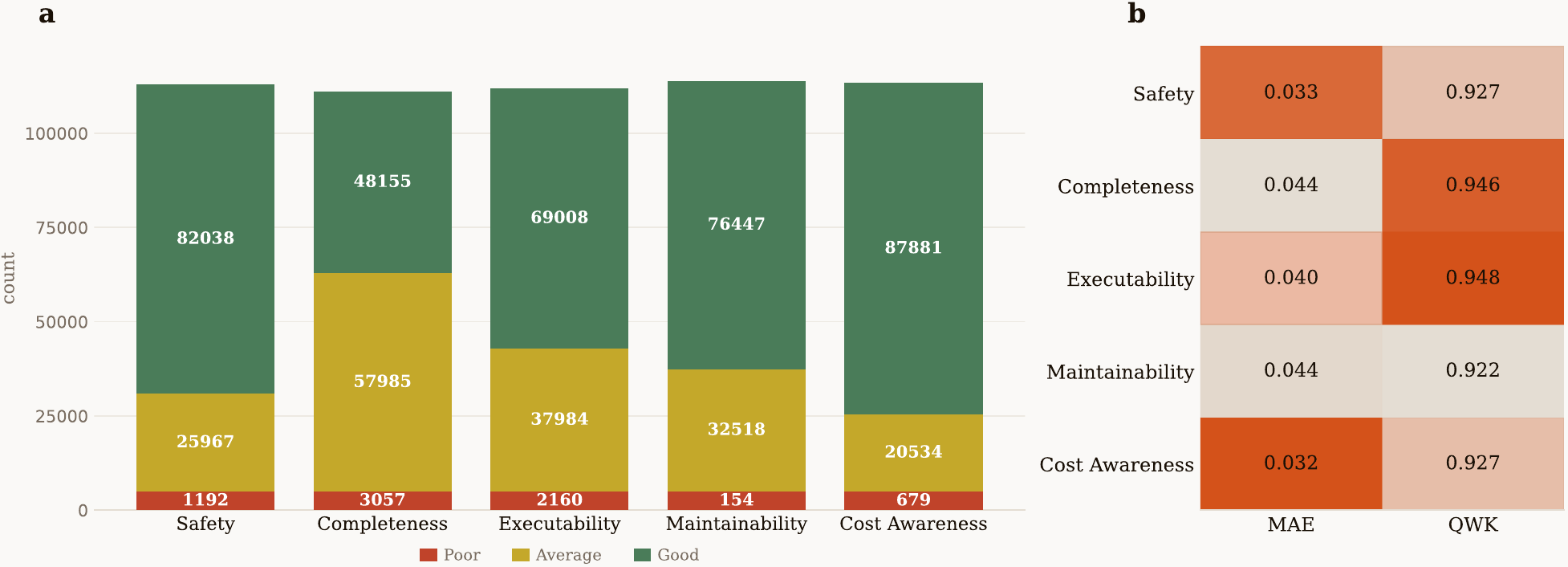}
    \caption{Multi-dimensional Skill Evaluation and Reliability Analysis.
    (a) Quality distribution of curated skills in SkillNet, assessed using a three-level grading scheme, where most skills are rated \textit{Good} or \textit{Average}.
    (b) Reliability validation of the automated evaluator on 200 randomly sampled skills. The heatmap shows mean absolute error (MAE) and quadratic weighted kappa (QWK) between human annotators and the model evaluator, with near-perfect QWK and low MAE across all dimensions, demonstrating the robustness and scalability of the evaluation framework.}
  \label{fig:eval_results}
\end{figure}
\item \textbf{Executability:} Verifies if the skill can be successfully implemented by agents in sandboxed environments, identifying hallucinated tool calls or ambiguous instructions.

\item \textbf{Maintainability:} Measures the modularity and composability of skills, ensuring they can be locally updated without disrupting global dependencies or breaking backward compatibility.

\item \textbf{Cost-awareness:} Quantifies execution overhead, including time latency, computational resource consumption, and API usage costs to support efficiency optimization.
\end{itemize}

To achieve high-throughput assessment, all dimensions are primarily evaluated via an automated LLM-based evaluator, guided by fine-grained rubrics. 
For \textbf{Safety}, the evaluator assesses potentially risky instructions and operations at the semantic level, providing a scalable repository-level signal for skill curation. 
This assessment is intended to support large-scale quality control rather than substitute for deployment-level security validation.
For \textbf{Executability}, we complement LLM judgment with empirical validation: skills containing code or tool invocations are executed in controlled sandbox environments to verify runtime correctness. 
Each dimension is categorized into three levels: \textit{Good}, \textit{Average}, and \textit{Poor}.

The quality distribution across the repository is illustrated in Figure \ref{fig:eval_results}(a). 
We observe that while \textbf{Safety} and \textbf{Maintainability} maintain a high proportion of ``Good'' ratings, \textbf{Executability} presents a greater challenge, with a larger fraction of skills rated as \textit{Average}. 
This reflects our rigorous filtering criteria, ensuring that only high-fidelity skills are prioritized for complex task execution.

To validate the reliability of this automated pipeline, we randomly sampled 200 skills and recruited three PhD-level computer science annotators for independent blind reviews. 
As shown in Figure \ref{fig:eval_results}(b), the inter-rater agreement between human judgments and LLM-based scores exhibits exceptional consistency. 
Across all dimensions, the Mean Absolute Error (MAE) remains below 0.03, and the Quadratic Weighted Kappa (QWK) consistently reaches near-perfect levels (above 0.90). These results confirm that our automated evaluator provides a human-aligned, robust, and scalable foundation for managing the SkillNet ecosystem.

\subsection{Skill Analysis}
Beyond isolated skill creation and evaluation, large-scale skill repositories introduce a new challenge: how to systematically understand, organize, and exploit the relations among skills. To address this, SkillNet introduces a dedicated Skill Analysis module that automatically discovers and models structural relations between skills, forming a structured and interpretable skill relation graph. This enables global reasoning over large skill repositories and supports advanced downstream applications such as skill retrieval, composition, dependency resolution, and workflow synthesis.

\paragraph{Relation Modeling.}
SkillNet formulates skill analysis as a structured relations discovery problem. Given a large amount of heterogeneous skills, the system automatically identifies and annotates multiple types of
\begin{figure}[h]
    \centering
    \includegraphics[width=\textwidth]{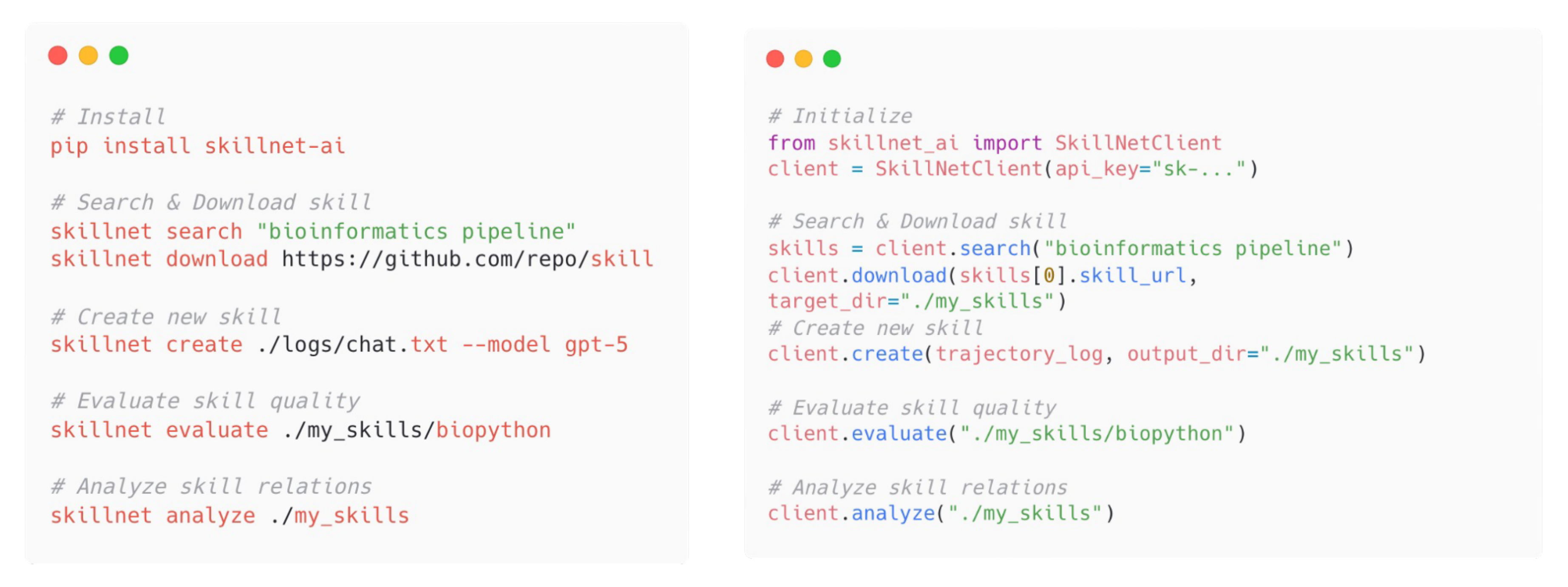} 
    \caption{Practical use examples of skillnet-ai. The package offers a unified functional experience through both a command-line interface (left) for interactive usage and a Python library (right) for seamless integration into AI development pipelines.}
    \label{fig:code_use}
\end{figure}
\noindent semantic and functional relations between skills, including (task-specific relations are omitted here, and users may extend the schema by defining custom relations as needed):
\begin{itemize}
    \item \textbf{similar\_to:} Two skills perform functionally equivalent or highly similar tasks and can often be used interchangeably, enabling redundancy detection, replacement, and robustness enhancement.

    \item \textbf{belong\_to:} A skill serves as a sub-component or atomic step within a larger composite workflow, capturing hierarchical structures and supporting skill abstraction and modularization.
    \item \textbf{compose\_with:} Two skills are frequently co-invoked in workflows, where one typically produces outputs consumed by the other, enabling automatic workflow composition and pipeline generation.
    \item \textbf{depend\_on:} A skill cannot execute independently without prerequisite skills, such as environment setup or API initialization, enabling explicit dependency tracking and safe execution planning.
\end{itemize}

\paragraph{Automated Skill Relation Graph Construction.}
SkillNet can construct large-scale skill relation graphs through a hybrid pipeline that integrates semantic embedding and LLM-based relational inference. 
Candidate relations are first generated via similarity matching, dependency extraction, and execution trace alignment, and LLM reasoning, enabling scalable and accurate discovery of structured skill relations.
These relations jointly form a directed, typed multi-relational graph, where nodes represent skills and edges encode fine-grained structural dependencies and functional associations.

% \paragraph{Applications and Capabilities.}
% The resulting skill graph enables a wide range of downstream capabilities. It supports fine-grained skill search and recommendation by exploiting semantic similarity and structural proximity. It enables automatic multi-skill composition by identifying compatible skill chains and resolving dependencies. It facilitates hierarchical task decomposition by traversing \texttt{belong\_to} relations, and enhances execution reliability through explicit dependency tracking. Moreover, the skill graph provides an interpretable substrate for analyzing capability coverage, redundancy, and evolution patterns within large agent ecosystems.

\paragraph{Task-Oriented Skill Collection Release.}
To promote reproducibility and practical adoption, SkillNet releases over 20 task-specific skill collections, covering domains such as scientific research automation, software development, data processing pipelines, web automation, etc. In particular, several collections are designed for academic benchmarks, and we conduct quantitative evaluations to assess the effectiveness of this paradigm, as detailed in a later section. 
These released collections provide structured and reusable infrastructures for studying large-scale agent planning, workflow synthesis, and skill evolution, offering a foundation for future research on structured agent intelligence.

\subsection{Open Resources}

SkillNet provides a comprehensive open infrastructure for creating, evaluating, and organizing AI skills at scale. This includes a large-scale skill repository, a front-end website, an open access API, and a versatile Python toolkit (skillnet-ai), forming a unified ecosystem for skill management and utilization.

The front-end website allows users to seamlessly browse, search, and download skills directly. Users can explore skill categories, view detailed API documentation and Python toolkit usage guidelines, and learn how to apply SkillNet to scientific research or coding scenarios. The platform also supports community
\begin{table*}[ht]
\centering
\small
\setlength{\tabcolsep}{6pt}
\renewcommand{\arraystretch}{1.2}
\begin{tabular}{l l cc cc cc cc cc}
\toprule
\multirow{3}{*}{Model} & \multirow{3}{*}{Method} 
& \multicolumn{4}{c}{ALFWorld} 
& \multicolumn{2}{c}{WebShop} 
& \multicolumn{4}{c}{ScienceWorld} \\
\cmidrule(lr){3-6} \cmidrule(lr){7-8} \cmidrule(lr){9-12}

& 
& \multicolumn{2}{c}{Seen} 
& \multicolumn{2}{c}{Unseen}
& \multirow{2}{*}{R ↑} & \multirow{2}{*}{S ↓} 
& \multicolumn{2}{c}{Seen}
& \multicolumn{2}{c}{Unseen} \\

\cmidrule(lr){3-4} \cmidrule(lr){5-6}
\cmidrule(lr){9-10} \cmidrule(lr){11-12}

& & R ↑ & S ↓ & R ↑ & S ↓ & & & R ↑ & S ↓ & R ↑ & S ↓ \\
\midrule

\multirow{3}{*}{DeepSeek V3.2} 
& React & 66.43 & 19.51 & 69.40 & 19.27 & \underline{31.55} & 24.06 & 69.86 & 17.59 & 64.67 & 19.26 \\
& Expel & \underline{67.86} & \underline{18.86} & \underline{76.12} & \underline{17.41} & 29.23 & \underline{24.00} & \underline{74.91} & \underline{15.98} & \underline{74.09} & \underline{17.53} \\
& + SkillNet & \textbf{80.60} & \textbf{14.54} & \textbf{83.57} & \textbf{14.81} & \textbf{46.18} & \textbf{17.84} & \textbf{84.87} & \textbf{11.89} & \textbf{81.31} & \textbf{12.48} \\
\midrule

\multirow{3}{*}{Gemini 2.5 Pro} 
& React & 60.00 & 18.72 & 61.94 & 19.18 & 31.66 & 22.12 & 58.24 & 18.42 & 56.13 & 19.07 \\
& Expel & \underline{68.57} & \underline{17.88} & \underline{70.15} & \underline{17.04} & \underline{33.12} & \underline{19.31} & \underline{72.76} & \underline{15.01} & \underline{67.37} & \underline{14.91} \\
& + SkillNet & \textbf{91.43} & \textbf{12.80} & \textbf{91.04} & \textbf{11.96} & \textbf{53.02} & \textbf{14.91} & \textbf{88.84} & \textbf{11.49} & \textbf{86.26} & \textbf{11.30} \\
\midrule

\multirow{3}{*}{o4 Mini} 
& React & 45.71 & 23.31 & 49.25 & 23.33 & 24.19 & 22.02 & 64.89 & 15.12 & 59.93 & 14.99 \\
& Expel & \underline{56.43} & \underline{21.35} & \underline{58.96} & \underline{21.85} & \underline{26.71} & \underline{21.91} & \underline{67.95} & \underline{13.65} & \underline{65.68} & \underline{13.95} \\
& + SkillNet & \textbf{68.57} & \textbf{18.94} & \textbf{73.28} & \textbf{17.08} & \textbf{36.21} & \textbf{18.79} & \textbf{73.24} & \textbf{13.30} & \textbf{71.06} & \textbf{12.35} \\
\midrule

\multirow{3}{*}{Claude Sonnet 4.6} 
& React & 88.57 & 12.00 & 89.55 & 12.32 & 41.09 & 21.58 & \underline{79.90} & 13.72 & \underline{78.16} & \underline{14.27} \\
& Expel & \underline{92.86} & \underline{11.82} & \underline{91.04} & \underline{12.29} & \underline{48.63} & \underline{20.07} & 75.99 & \underline{12.99} & 76.24 & 14.88 \\
& + SkillNet & \textbf{95.71} & \textbf{10.68} & \textbf{95.52} & \textbf{11.03} & \textbf{50.92} & \textbf{19.14} & \textbf{83.25} & \textbf{10.77} & \textbf{82.94} & \textbf{11.91} \\
\bottomrule
\end{tabular}

\caption{Experimental results on ALFWorld, WebShop, and ScienceWorld. 
R denotes average reward, S denotes average steps (↑ indicates
the larger values are better, and ↓ denotes the smaller values are better). The top-performing results are highlighted in \textbf{bold}, and the second-best are \underline{underlined}.}
\label{tab:quant_exp}
\end{table*}
\noindent contributions, enabling users to submit and share newly constructed skills, thus fostering a collaborative and continuously evolving skill ecosystem.

For API access, SkillNet supports both keyword-based and vector-based searches. Users can search for skills by keywords, categories, or semantic similarity via \url{http://api-skillnet.openkg.cn/v1/search}.

To bridge the gap between static repositories and dynamic execution, we developed \texttt{skillnet-ai}, a unified Python library and CLI tool that enables flexible integration with users. Through this toolkit, users can search for skills using keywords or semantic similarity, download skills directly from GitHub into local workspaces, and create structured skills from heterogeneous sources, including execution trajectories, GitHub repositories, office documents, and natural language prompts. The package also supports evaluating skills across multiple dimensions (safety, completeness, executability, maintainability, and cost-awareness) to ensure practical reliability, and analyzing relations between skills to uncover dependencies, hierarchical compositions, collaborations, and functional similarities. The practical use examples are illustrated in Figure~\ref{fig:code_use}.

SkillNet has initially aggregated over 600k candidate skills from open internet resources, automated creation pipelines, and community contributions. After multi-stage filtering and evaluating, more than 500k high-quality skills are curated in the final repository (and which is constantly expanding). 
This scale ensures broad coverage across scenarios while maintaining rigorous quality standards, enabling SkillNet to serve as a reliable skill infrastructure for both scientific investigation and real-world deployment.
We actively encourage community contributions, and all community-submitted skills are subject to the same automated quality checks. In addition, we conduct periodic manual audits through random sampling to ensure the ongoing reliability and integrity of the shared skill repository.

\section{Quantitative Evaluation}

\subsection{Settings}
To quantitatively assess the effectiveness of SkillNet, we conduct experiments across three text-based simulated environments. 
ALFWorld \cite{shridhar2021alfworldaligningtextembodied} provides an embodied household environment, requiring agents to navigate and manipulate objects to complete daily tasks; WebShop \cite{yao2023webshopscalablerealworldweb} simulates realistic online shopping scenarios, where agents perform product search, comparison, and purchasing under specified constraints; and ScienceWorld \cite{wang2022scienceworldagentsmarter5th} presents a virtual scientific laboratory in which agents must conduct experiments and operate scientific instruments. All three environments are formulated as partially observable Markov decision processes (POMDPs), requiring agents to make sequential decisions under incomplete and noisy observations.

\begin{figure}[ht]
    \centering
    \includegraphics[width=1\textwidth]{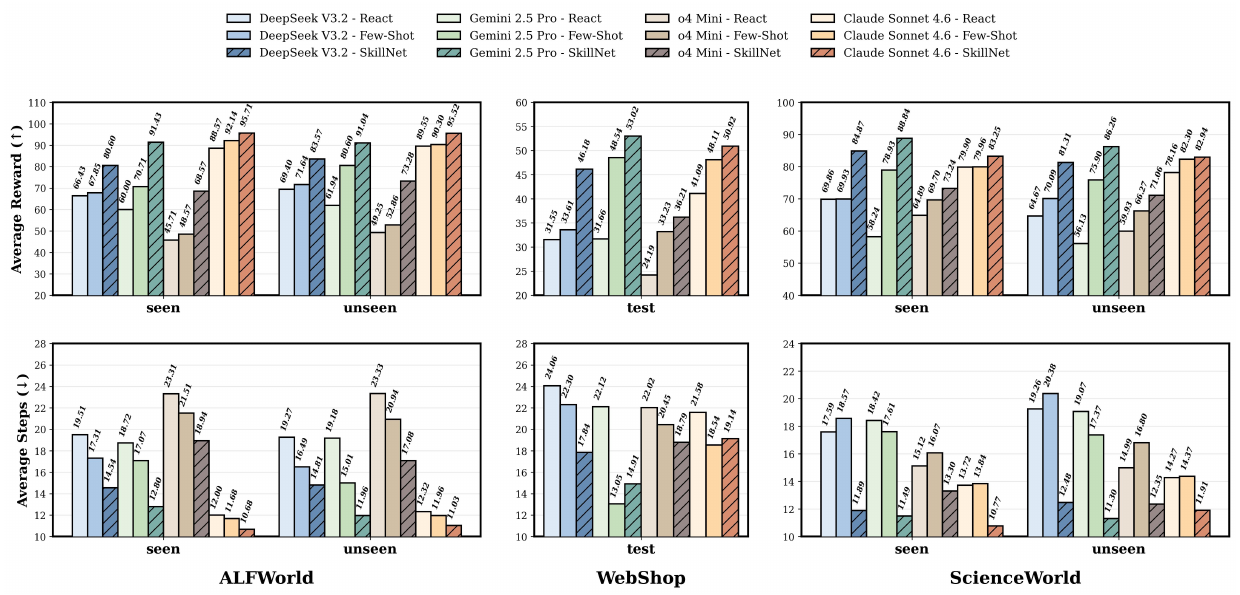} 
    \caption{Performance comparison across diverse methods and models. The results illustrate that SkillNet consistently outperforms React and Few-shot baselines, achieving significantly higher average rewards (top) and reduced average steps (bottom) across ALFWorld, WebShop, and ScienceWorld.}
    \label{fig:quant_exp}
\end{figure}

We adopt ReAct \cite{yao2023reactsynergizingreasoningacting}, Expel \cite{zhao2024expelllmagentsexperiential} and a standard Few-Shot approach as the baseline methods. 
ReAct interleaves reasoning, action, and observation for step-by-step task solving. 
ExpeL gathers experiences from past tasks to extract natural language insights and retrieves past successful trajectories as in-context examples during inference. 
The Few-Shot baseline randomly samples a complete expert trajectory to serve as a static in-context exemplar. 
In contrast, our proposed variant is augmented with SkillNet by leveraging expert trajectories from ETO \cite{song2024trialerrorexplorationbasedtrajectory} to synthesize benchmark-specific skill collections (these collections are integrated into SkillNet, and users can extend it to other datasets by creating corresponding skills)\footnote{\url{https://github.com/zjunlp/SkillNet/tree/main/experiments/src/skills}.}. 
Note that the above experiences and the seen and unseen splits of the test set have no overlap, thereby preventing data leakage.
During evaluation, agents are equipped with these collections and can dynamically select, activate, and execute the most relevant skills based on the current state. We employ four representative LLMs: DeepSeek V3.2 \cite{deepseekv3}, Gemini 2.5 Pro \cite{gemini2.5pro}, o4 Mini \cite{o4mini}, and Claude Sonnet 4.6 \cite{claude46} as backbone models.

\subsection{Results}

Table~\ref{tab:quant_exp} and Figure~\ref{fig:quant_exp} present the quantitative results across diverse environments and backbone models. In general, integrating SkillNet consistently yields substantial gains in both task effectiveness and execution efficiency, validating the effectiveness of our framework.
Compared with ReAct, SkillNet improves the average reward by 40\%, while reducing the number of interaction steps by 30\% on average, indicating that agents equipped with SkillNet are able to solve tasks more reliably while executing shorter and more coherent action trajectories. 
This reflects SkillNet’s ability to transform fragmented experience into reusable procedural abstractions, enabling agents to perform structured planning, reduce redundant exploration, and execute complex behaviors in a more principled and systematic manner.

Importantly, the performance improvements remain robust across backbone models of varying capacities, ranging from compact models (+15.7 R for o4 Mini) to large-scale LLMs (+28.5 R for Gemini 2.5 Pro). 
This demonstrates that SkillNet provides complementary capabilities beyond parametric knowledge, effectively augmenting reasoning with persistent, executable skills. 
Moreover, the consistent gains observed under both seen and unseen settings highlight SkillNet’s strong generalization ability, suggesting that skill abstraction and reuse facilitate knowledge transfer across tasks and environments.

Taken together, these results substantiate the central design principle of SkillNet: by formalizing skills as independent, systematically accumulated, knowledge-grounded capability units, agent competence can be enhanced cumulatively rather than episodically, ultimately fostering domain-specialized and continually self-improving agents.
\section{SkillNet-Gym}

\begin{figure*}[ht]
    \centering
    \includegraphics[width=1\textwidth]{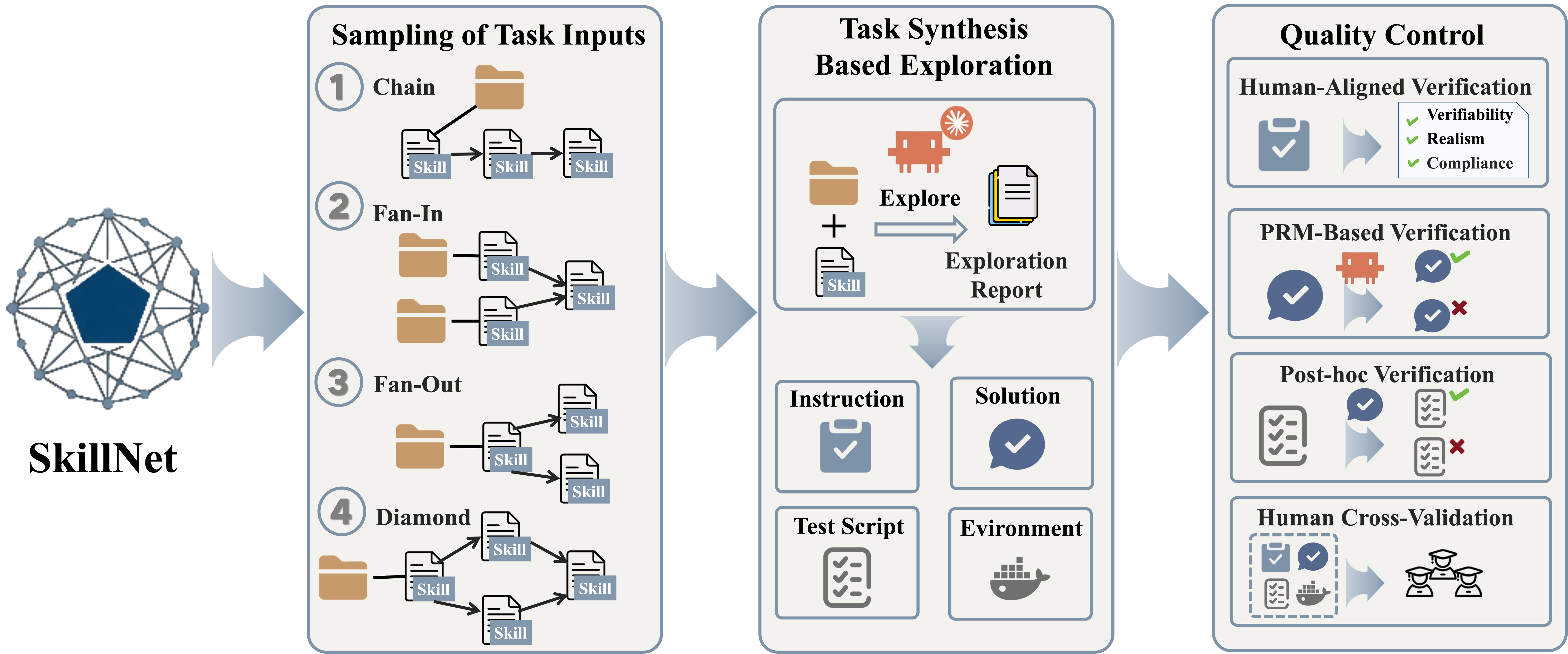}
    \caption{The overview of \dataname construction pipeline.}
    \label{fig:pipeline}
\end{figure*}

\subsection{Design Principles}
To reflect the demands of realistic and continuously evolving community, we design \dataname around four key properties: \emph{(1) Real-World Source:} \dataname is grounded in artifacts collected from real community resources, rather than manually curated skills or synthetic files. This design choice ensures that the benchmark better reflects the structure, variability, and messiness of real-world usage scenarios.
\emph{(2) Domain diversity:}
\dataname spans 13 core domains and 81 subdomains, covering a broad range of practical settings, including data analysis, science, technology and so on.
% , as illustrated in Figure~\ref{fig:taxonomy}.
\emph{(3) Dynamic extensibility:}
To accommodate the continuously evolving community ecosystem, SkillNet is represented as a heterogeneous graph that can be incrementally expanded by adding new artifact nodes and their associated edges. This formulation enables the benchmark to evolve over time.
\emph{(4) Automatic construction:} Our pipeline automatically constructs the benchmark from SkillNet at scale, and also supports automatic control over task difficulty by varying the number of nodes visited during random walks and the topological structures induced by the traversal.
The comparison of \dataname with other benchmarks is detailed in Table~\ref{tab:benchmark_comparison}.
Next, we introduce the automated pipeline of \dataname, covering SkillNet construction, benchmark data synthesis, and quality control. The pipeline is shown in Figure~\ref{fig:pipeline}.

\begin{table*}[h!]
\setlength{\tabcolsep}{3pt}
\centering
\resizebox{\linewidth}{!}{%
\begin{tabular}{@{}lcccccccc@{}}
    \toprule[1.5pt]
    \multirow{3}{*}{\textbf{Benchmark}} & \multicolumn{3}{c}{\textbf{Benchmark Construction}} & \multicolumn{3}{c}{\textbf{Benchmark Scope}} & \multicolumn{2}{c}{\textbf{Evaluation Focus}} \\ 
    \cmidrule(lr){2-4} \cmidrule(lr){5-7} \cmidrule(lr){8-9} 
    & \makecell{Real-World\\Source} & \makecell{Automatic\\Construction} & \makecell{Dynamic\\Extensibility} & \makecell{Domain\\Coverage} & \makecell{Skill\\Pool} & \makecell{Task\\Numbers} & \makecell{Skill\\Construction} & \makecell{Skill\\Composition} \\
    % & \makecell{Skill\\Efficacy}  \\ 
    \midrule
    % SkillCraft \cite{chen2026skillcraftllmagentslearn}   & \xmark & \xmark & \xmark & $28$ & \cmark & \xmark & { }{ }$[2,10]$ & \xmark & \xmark \\
    SkillsBench \cite{skillsbench2025} & \notcheckmark & \xmark & \xmark & \cmark & \xmark & $87$ & \xmark & \xmark \\
    % & \cmark \\
    SWE-Skills-Bench \cite{han2026sweskillsbenchagentskillsactually} & \notcheckmark & \xmark & \xmark & \xmark & \xmark & $565$ & \xmark & \xmark \\
    % & \cmark \\
    SkillLearnBench \cite{zhong2026skilllearnbenchbenchmarkingcontinuallearning} & \notcheckmark & \xmark & \xmark & \notcheckmark & \xmark & $20$ & \cmark & \xmark \\
    % & \cmark \\
    WildSkillBench \cite{liu2026agenticskillsworkwild}  & \notcheckmark & \xmark & \xmark & \cmark & \cmark & $87$ & \cmark & \cmark \\
    % & \cmark \\
    SkillRouter \citep{zheng2026skillrouterskillroutingllm}  & \cmark & \cmark & \cmark & \cmark & \cmark & $75$ & \xmark & \cmark \\
    % & \xmark  \\
    SkillRet \citep{cho2026skillretlargescalebenchmarkskill}  & \cmark & \cmark & \cmark & \cmark & \cmark & $4997$ & \xmark & \cmark \\
    % & \xmark \\
    AgentSkillOS \cite{li2026organizingorchestratingbenchmarkingagent}  & \cmark & \xmark & \xmark & \notcheckmark & \cmark & $30$ & \xmark & \cmark \\
    % & \cmark  \\
    SkillGenBench \cite{zhou2026skillgenbenchbenchmarkingskillgeneration}  & \cmark & \cmark & \cmark & \notcheckmark & \xmark & $187$ & \cmark & \xmark \\
    % & \cmark   \\
    OpenSkillEval \cite{ying2026openskillevalautomaticallyauditingopen}  & \cmark & \cmark & \cmark & \notcheckmark & \xmark & $600$ & \xmark & \xmark \\
    % & \cmark   \\
    \midrule[0.3pt]
    \dataname (ours) & \cmark & \cmark & \cmark & \cmark & \cmark & $159$ & \cmark & \cmark \\
    % & \cmark \\
    \bottomrule[1.5pt]
\end{tabular}}
\caption{Comparison of existing agentic skills benchmarks across three  dimensions: \textbf{Benchmark Construction}, \textbf{Benchmark Scope} and \textbf{Evaluation Focus}. ``{\cmark}''~indicates fully coverage, ``{\notcheckmark}''~indicates partially coverage, and ``{\xmark}''~indicates not coverage. Detailed explanations for each trait are provided in Appendix~\ref{appdix:comparison}.}
\label{tab:benchmark_comparison}
\end{table*}

\subsection{Skill Graph Construction}
We construct skill graph (customized SkillNet) from real community resources by collecting three aligned artifact types: skills, supporting documents, and processable files.
Specifically, we retrieve candidate skills from repositories such as ClawHub and SkillHub using LLM-generated domain-diverse queries, then perform filtering and deduplication.
For each skill, we gather web documentation based on its functional semantics, or synthesize surrogate documentation when necessary, and retain functionally compatible files with source metadata for downstream normalization and quality control. On top of these artifacts, we build SkillNet as a skill-centric DAG: an LLM extracts each skill’s pre and post scenarios, we identify candidate edges by matching embedded post-scenarios to pre-scenarios, and then use an LLM to verify whether each match constitutes a valid workflow handoff. 
The resulting graph captures directed compositional dependencies among skills and, compared with scenario-centric formulations, more directly models skill topology for multi-skill task synthesis.

\subsection{Automatic Task Synthesis from SkillNet}

\paragraph{Diversity Sampling of Task Inputs.}
Given the constructed SkillNet, we synthesize task candidates by sampling structured subgraphs from the directed skill graph. We consider four common composition patterns: chains, fan-out, fan-in, and diamonds, with two to five skills depending on the structure, as illustrated in Figure~\ref{fig:task-topology-patterns}. Each candidate preserves its participating skills, directed skill connections, and scenario-level evidence for every edge. We score each edge using alignment confidence, retrieval similarity, and the number of supporting scenario alignments, and score each task by combining mean edge quality with a structure bonus and a hub-node penalty. The resulting candidates are grouped by structure category and assigned difficulty labels from easy to expert according to their compositional complexity. We then select tasks either by quality ranking. This produces a pool of compositional task candidates grounded in verified SkillNet transitions.

\paragraph{Agent-Driven Exploration Report Generation.}
Given the sampled skill Directed Acyclic Graph (DAG) and its associated input artifacts, we prompt an agent (Claude Code) to execute the skills by following the workflow specified by the DAG and operating directly on the input files. 
After the exploration phase, the agent produces an exploration report that summarizes actionable findings, such as feasible cross-skill workflows, potential intermediate artifacts, and common failure patterns. 
Compared with naively concatenating the full contents of all skill files, the exploration report is substantially more concise and better aligned with downstream task requirements.

\paragraph{Task Synthesis.}
A complete task should include at least four core components: a task instruction, a golden solution, an evaluation script, and the execution environment. We synthesize these components sequentially.
\textit{1) Task Instruction}: Given the exploration report and snippet-level information about the input artifacts, we prompt an LLM to generate a task instruction that follows the workflow induced by the skill DAG. To encourage diversity, each synthesis run uses a randomly sampled writing style.
\textit{2) Golden Solution}: After generating the instruction, we first derive high-level guidance from the exploration report to scaffold the downstream problem-solving process. We then prompt Claude Code to solve the task conditioned on this meta-information. From the resulting execution trajectory, we extract and compress the essential steps into a canonical reference solution, \texttt{solve.sh}.
\textit{3) Test Script}: Based on the task instruction and the final output artifacts, we synthesize a test script \texttt{test.sh} that verifies not only the structural validity of the outputs but also their semantic correctness at the value level.
\textit{4) Execution Environment}: Finally, we extract the minimal execution environment to get a \texttt{Dockerfile} from \texttt{solve.sh}.

\subsection{Quality Control}
We apply rigorous multi-stage quality control to the benchmark. 
First, during SkillNet construction, we filter low-quality nodes and edges to improve the reliability of the underlying skill graph. 
Second, for synthesized tasks, task-level quality control covers four core components of each instance: the instruction, the reference

\begin{table*}[ht!]
    \centering
    \newcolumntype{C}{>{\centering\arraybackslash}p{1.1cm}}
    \resizebox{0.99\textwidth}{!}{%
    % \begin{tabular}{l|*{10}{C}}
    \begin{tabular}{l|CC|CC|CC|CC|CC}

    \toprule
    \multirow{3}{*}{\textbf{Models}}
    & \multicolumn{6}{c|}{\textit{Skill Construction}}
    & \multicolumn{4}{c}{\textit{Skill Composition}} \\
    \cmidrule(lr){2-7} \cmidrule(lr){8-11}
    & \multicolumn{2}{c|}{\textit{No Skill}}
    & \multicolumn{2}{c|}{\textit{Official Skill}}
    & \multicolumn{2}{c|}{\textit{Skill-Creator}}
    & \multicolumn{2}{c|}{\textit{Skills Retrieve}}
    & \multicolumn{2}{c}{\textit{Skills Orchestration}} \\
    \cmidrule(lr){2-3} \cmidrule(lr){4-5} \cmidrule(lr){6-7} \cmidrule(lr){8-9} \cmidrule(lr){10-11}
    & \textit{Easy} & \textit{Hard}
    & \textit{Easy} & \textit{Hard}
    & \textit{Easy} & \textit{Hard}
    & \textit{Easy} & \textit{Hard}
    & \textit{Easy} & \textit{Hard} \\
    \midrule

    \rowcolor{mydarkbrown!40}
    \multicolumn{11}{c}{\textit{\textbf{Claude Code}}} \\

    Minimax-M2.7 & 61.04 & 17.89 & 65.58 & 26.22 & 66.88 & 20.73 & 8.12 & 13.10 & 18.55 & 11.62\\
    Kimi-K2.6 & 70.13 & 26.83 & 74.03 & 34.15 & 73.38 & \underline{31.10} & 14.29 & 19.51 & 4.62 & 12.66 \\

    Deepseek-V4 & 65.28 & 23.38 & 69.44 & 29.87 & 68.83 & 29.88 & \underline{24.68} & \underline{30.49} & \textbf{22.22} & \underline{24.36}\\
    GLM-5.1 & 70.13 & 26.83 & 71.43 & 32.93 & \textbf{79.22} & \textbf{31.10} & 16.02 & 25.20 & 8.72 & 20.68 \\

    Claude 4.5 Haiku & 59.09 & 16.46 & 55.84 & 26.22 & 53.90 & 23.17 & 15.21 & 21.65 & 8.72 & 9.33 \\
    Claude 4.6 Sonnet & 62.34 & 20.73 & 74.03 & \underline{36.59} & 72.73 & 30.49 & \textbf{25.98} & \textbf{39.17} & \underline{14.59} & \textbf{26.70} \\

    \midrule
    \rowcolor{sky!40}
    \multicolumn{11}{c}{\textit{\textbf{Codex}}} \\
    \midrule
    GPT-5.4 mini & \underline{72.73} & \underline{26.83} & \underline{74.03} & 32.93 & \underline{75.32} & 30.49 & 5.63 & 14.63 & 6.67 & 15.19 \\
    GPT-5.4 & \textbf{74.68} & \textbf{28.66} & \textbf{75.32} & \textbf{37.20} & 74.03 & 26.83 & 11.26 & 18.29 & 11.35 & 16.46 \\

    \bottomrule
    \end{tabular}%
    }
\caption{Comparison across different harnesses and models under different skill usage settings. 
\textbf{No Skill} uses no external skills;
\textbf{Official Skill} directly provides the task-associated official skills;
\textbf{Skill-Creator} evaluates execution with skills constructed from upstream resources;
\textbf{Skills Retrieve} requires agents to retrieve relevant skills from a large wild skill pool before execution;
\textbf{Skills Orchestration} additionally requires dependency-aware organization of the selected skills into an execution workflow.
Best and second-best results within each column are shown in \textbf{bold} and underlined, respectively.
}
\label{tab:all_databases}
\end{table*}

\noindent solution, the test script, and the task difficulty. 
Through this rigorous quality control pipeline, we ultimately retain 159 high-quality tasks.
Detailed quality control procedures are provided in Appendix~\ref{appdix:quality}.

\subsection{Difficulty Stratification for Analysis}
For analysis, we further partition the tasks into two coarse difficulty strata, \textbf{Easy} (77 tasks) and \textbf{Hard} (82 tasks), using post hoc empirical performance as a calibration signal while maintaining balanced domain coverage across the two subsets. This stratification enables more fine-grained comparisons of model behavior across tasks with different realized levels of difficulty.

\subsection{Evaluation Methodology and Metrics}

\paragraph{Skill Construction.}
We study agent behavior from the perspective of skill lifecycle:
\emph{(1) No Skill:} In the initial stage, the agent solves tasks without access to any skills.
\emph{(2) Official Skill:} The agent can leverage externally provided skills, such as official skills released in community repositories like SkillHub and ClawHub.
\emph{(3) Skill-Creator:} Using meta-skill, e.g., Skill-Creator, the model distills human-authored procedural documents into reusable skills that it can invoke in future task execution.
In the final setting, the agent executes tasks end-to-end under both settings: with skills and without skills.
For evaluation, each execution trajectory receives a reward of $1$ if the agent’s final output passes all unit tests, and $0$ otherwise. Each task is attempted independently for $k$ runs, and we report \textbf{Avg@$k$}.

\paragraph{Skill Composition}
Beyond end-to-end evaluation, we also consider a finer-grained setting designed to assess how an agent retrieves and orchestrates skills from a realistic skill pool to solve a given task. 
In this setting, the agent is not required to fully execute each task. Instead, we focus on its ability to identify appropriate skills from the skill pool, which is constructed from the 2,000+ skills included in SkillNet 
% (see Appendix~\ref{appdix:skillnet_statics} for details) 
and organize them into an effective execution workflow.
We report two stage-specific completeness metrics: (1) Skill Retrieval Completeness, which measures whether the retrieved skill nodes fully cover the gold skill set; (2) Skill Orchestration Completeness, which measures whether the predicted dependency edges fully recover the gold workflow among retrieved skills.
Both metrics use binary completeness because our goal is to evaluate whether an agent can recover a complete executable composition rather than partial overlap.

\subsection{Experimental settings}

\paragraph{Agent Harnesses.}
We evaluate the two most popular agents in current use: Claude Code~\citep{anthropic2025claudecode} and Codex CLI~\citep{openai2025codexcli}.

\paragraph{Models.}
We evaluate eight frontier models, including four open-source models and four proprietary models: MiniMax-M2.7 \cite{minimax2series}, Kimi-K2.6 \cite{kimiteam2026kimik25visualagentic}, Deepseek-V4 \cite{deepseekai2026deepseekv4}, GLM-5.1 \cite{glm5team2026glm5vibecodingagentic}, Claude Haiku 4.5, Claude Sonnet 4.6, GPT-5.3 Codex, and GPT-5.4. 
All models are evaluated with their default temperature settings, and each task is independently run three times. 
Following commonly adopted community practices, we pair the open-source models and the Claude series with Claude Code, while the GPT series is paired with Codex.

\subsection{Main Results}

% 执行 构建 检索 编排
\paragraph{Skills Help Hard Tasks More Reliably Than Easy Tasks.}
Compared with the easy tasks, skill augmentation generally yields larger and more consistent relative improvements on the hard tasks. 
A likely reason is that current frontier models already possess strong agentic capabilities, even without external skills, they can often solve many easy tasks reasonably well. 
Consequently, the headroom for improvement on easy tasks is limited, whereas challenging tasks benefit more substantially from explicit skill support. The corresponding results are reported in Table~\ref{tab:all_databases}.

\paragraph{One-Shot Skill Construction May Introduces Information Loss.}
Although the model is provided with documents of higher information density, compressing them into a skill inevitably discards part of the original information, which can lead to degraded performance for tasks requiring specific capabilities. The corresponding evidence is shown in Table~\ref{tab:all_databases}. 
At the same time, our results also suggest that officially curated skills are not always superior to self-constructed ones, highlighting the importance of model-side adaptation. 
In addition, we evaluate another skill-construction baseline, SkillSeeker \cite{skillseekers}, whose results are reported in Table~\ref{tab:construction_baseline} and exhibit similar trends. 
While our current study only considers a single-pass compression strategy, existing skill-construction methods are increasingly moving toward more sophisticated pipelines \cite{chen2026skvmrevisitinglanguagevm,zhang2026coevoskillsselfevolvingagentskills,si2026contextskillslanguagemodels} that enable adaptation and self-evolution. We believe that, in the future, model constructed skills may eventually surpass currently human curated skills.

\paragraph{Frontier Agents Exhibit Weak Skill Composition Ability in the Wild.}
Although frontier agents can benefit from provided skills in controlled settings, their ability to operate over large, community-sourced skill repositories remains limited. Table~\ref{tab:all_databases} reveals a notable difficulty inversion between skill retrieval and skill orchestration. Hard tasks often obtain higher retrieval scores than easy tasks, suggesting that complex instructions expose richer procedural cues and make the required skills more identifiable. However, this advantage does not translate into successful orchestration. Once the agent must organize the retrieved skills into a dependency-preserving workflow, hard tasks become substantially more challenging.
We also evaluate end-to-end execution in a realistic wild setting in Table~\ref{tab:in_wild}.
Compared with directly providing official skills, exposing agents to a large skill library leads to clear performance degradation, with the drop being more pronounced on hard tasks. These results show that incomplete retrieval and incorrect orchestration are not only intermediate failures, but directly limit downstream task success. Therefore, effective real-world skill use requires agents to go beyond consuming fixed skills and develop adaptive orchestration mechanisms that can expand exploration, verify dependencies, and update skill workflows as the skill ecosystem evolves.

\begin{table}[htbp]
\centering

\begin{minipage}{0.45\textwidth}
\centering
\footnotesize
\renewcommand{\arraystretch}{1.05}
\setlength{\tabcolsep}{2pt}

\begin{tabular}{@{}lcccc@{}}
\toprule
\textbf{Methods} & \textbf{Complete} & \textbf{Recall} & \textbf{Precision} & \textbf{Avg sel.} \\
\midrule

\multicolumn{5}{c}{\textbf{Skills Retrieve}} \\ 
\midrule

Agent & 18.83 & 44.93 & 37.92 & 3.86\\
AgentSkillOS & 42.14 & 61.77 & \textbf{55.00} & -\\
SkillRouter & & & & \\
~~~w/o Reranker & 36.48 & 60.28 & 19.06 & -\\
~~~w/ Reranker & 44.03 & 67.40 & 21.01 & -\\
SkillRet & \textbf{59.12} & \textbf{78.40} & 24.09 & -\\

\midrule

\multicolumn{5}{c}{\textbf{Skills Orchestration}} \\ 
\midrule

Agent & 18.36 & 25.17 & 18.63 & 3.27 \\
AgentSkillOS & \textbf{29.49} & \textbf{35.51} & \textbf{26.86} & - \\

\bottomrule
\end{tabular}

\captionof{table}{Performance of Skill Composition Baselines.}
\label{tab:final_compact_matrix_clean}

\end{minipage}
\hfill
\begin{minipage}{0.48\textwidth}
\centering

\includegraphics[width=\textwidth]{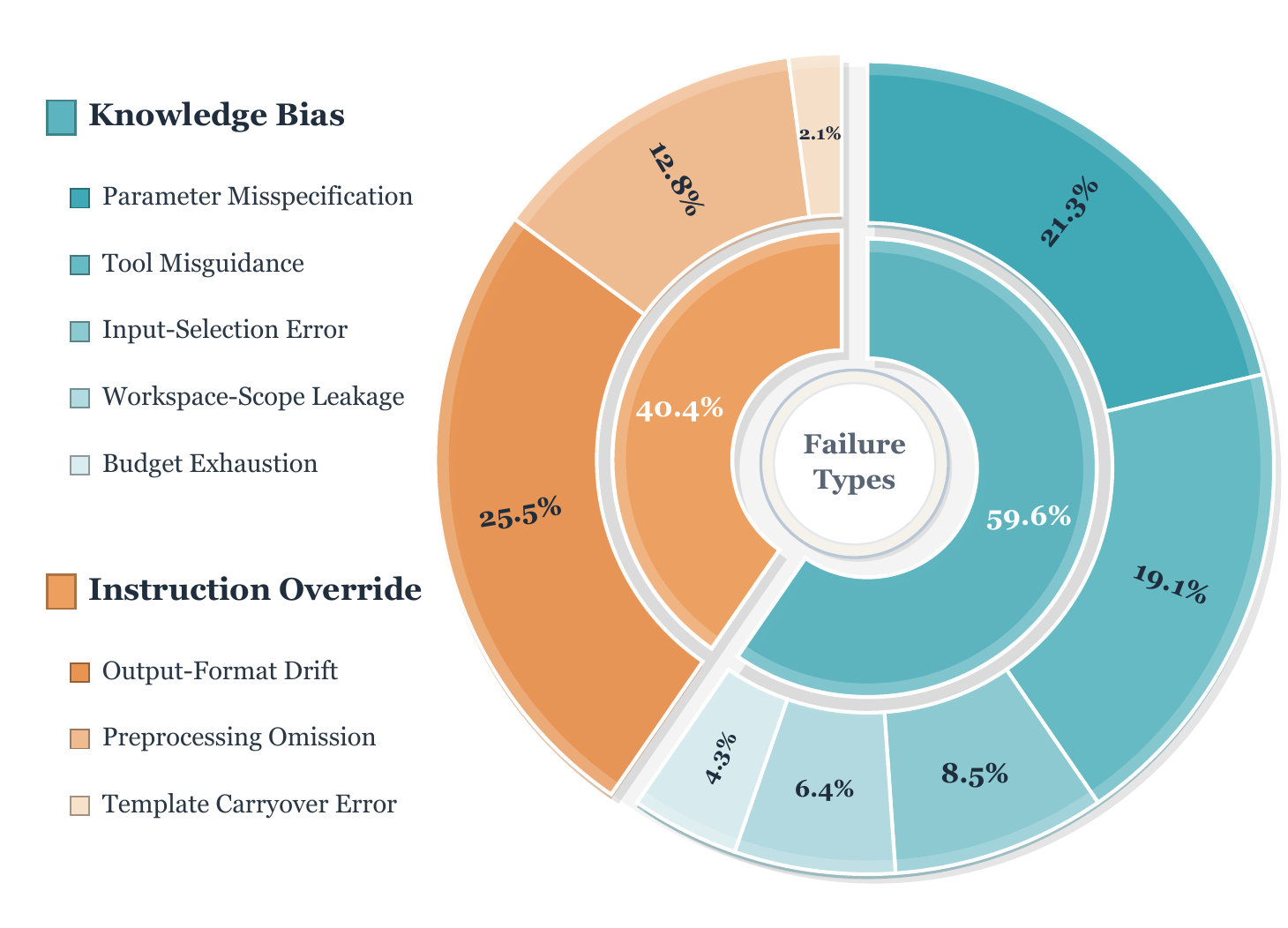}

\captionof{figure}{Failure Types}
\label{fig:failure_types}

\end{minipage}

\end{table}

\subsection{Analysis}

\paragraph{Specialized Composition Baselines Remain Stronger than Agents.}
We further evaluate representative baselines specifically designed for skill composition, including SkillRouter \cite{zheng2026skillrouterskillroutingllm}, SkillRet \cite{cho2026skillretlargescalebenchmarkskill}, and AgentSkillOS \cite{li2026organizingorchestratingbenchmarkingagent}, and compare them against the average performance of the eight agents on Top-10. 
The results are shown in Table~\ref{tab:final_compact_matrix_clean}.
Here, Avg. sel denotes the average number of skills retrieved and the average number of dependency edges selected by the agent. 
Among the retrieval baselines, SkillRet achieves substantially higher recall than SkillRouter, despite the latter using a reranker, likely due to SkillRet's access to larger-scale training data. For AgentSkillOS, we do not adopt its original high-cost tree-construction pipeline; instead, we prompt GLM-5.1 to perform selection and orchestration based only on the preliminary candidates returned by SkillRet. 
AgentSkillOS substantially outperforms direct skill orchestration from the skill repository. These results highlight the importance of hybrid retrieval and orchestration strategies for robust skill composition in the wild.

\begin{figure*}[t]
    \centering
    % \vspace{3mm}
    \includegraphics[width=1\linewidth]{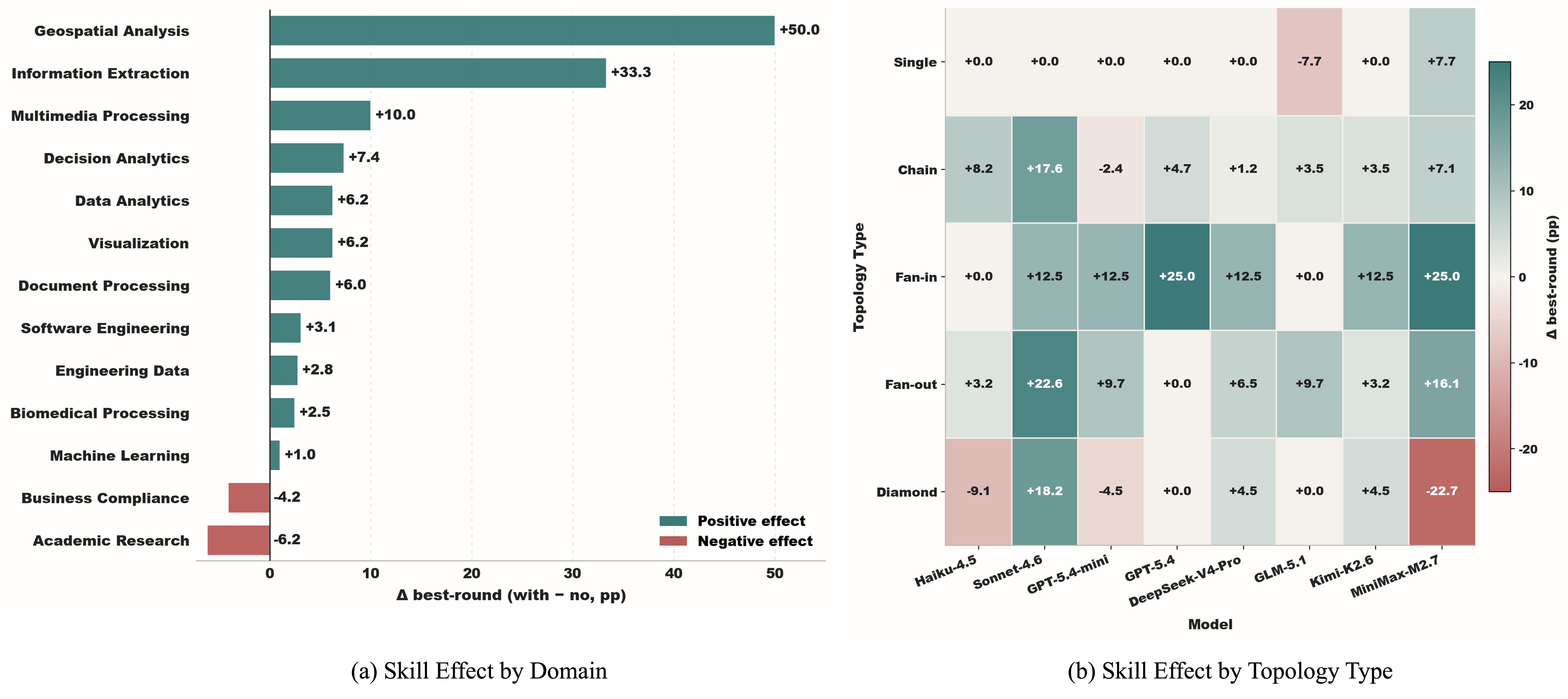}
    \vspace{-9mm}
    \caption{
    % \textbf{Comprehensive Analysis of \dataname.}
    \textbf{Skill utility across domains and task topologies.}
    \textbf{(a) Domain-wise Skill Effect:} The bar chart reports the average performance delta of skill augmentation over the no skill setting.
    \textbf{(b) Topology-wise skill effect:} The heatmap shows the mean delta for each topology across eight models, revealing that skill effectiveness depends jointly on task structure and model capability.
}
    % \vspace{-2mm}
    \label{fig:analysis}
\end{figure*}

\paragraph{Analysis of Failure Types.}
To analyze bad cases, defined as tasks for which injecting a skill strictly reduces agent accuracy, an initial round of manual inspection is conducted and then extended with AI-assisted analysis. The resulting failures are organized into two level categories (Figure~\ref{fig:failure_types}(c)).
\emph{(1) Knowledge Bias (59.6\%) .} This category covers cases in which the injected skill provides incorrect or misleading content. Representative patterns include \emph{tool misguidance}, such as recommending an inappropriate library, version, or API; \emph{parameter misspecification}, such as using an incorrect threshold, regular expression, or comparison direction; and \emph{input selection errors}, where the wrong record or sub-table is selected under the skill’s decision rule. Related failures also arise from \emph{workspace-scope leakage}, where hard-coded paths in the skill trigger reads or writes outside the target task workspace, and \emph{budget exhaustion}, where extra subroutines introduced by the skill consume the step or token budget and ultimately truncate execution. These results suggest that \textbf{skill construction should be adaptive, so as to reduce the mismatch between human curated skills and the model's actual execution behavior.
However, effective self-adaptation typically requires a dynamic evaluation benchmark to guide the direction of optimization. In this regard, \dataname can serve as a particularly useful testbed.}
\emph{(2) Instruction Override (40.4\%) .} In this failure mode, the injected skill overrides requirements explicitly specified in the task prompt. This typically manifests as \emph{output-format drift}, where the generated result follows an incorrect JSON schema, data type, or sign convention; \emph{preprocessing omission}, where a required normalization or editing step is skipped; and \emph{template carryover}, where scaffolding introduced by the skill is only partially completed. This pattern suggests that \textbf{current foundation models still struggle to balance contextual knowledge against direct instruction following.} More capable next generation models may need to internalize skill utilization as a more native capability, rather than treating skills as external additions that can conflict with task instructions.

% \paragraph{Finding 7: taxonomy analysis}
\paragraph{Skill Benefits Are Highly Domain Dependent.}
Skill utility varies considerably across domains (Figure~\ref{fig:analysis}(a)). Tasks with highly procedural workflows, such as Geospatial Analysis (+50.0 pp) and Information Extraction (+33.3 pp), benefit the most, suggesting that reusable execution patterns can be effectively encapsulated into skills. Moderate gains are observed in analytics and visualization tasks, where skills mainly reduce tool use overhead while leaving substantial reasoning to the model. In contrast, Academic Research (-6.2 pp) and Business Compliance (-4.2 pp) exhibit slight performance degradation. These domains typically require open-ended exploration, adaptive planning, and dynamic evidence integration, where rigid procedural skills may prematurely constrain the agent's search strategy. Overall, the results indicate that the effectiveness of skills is strongly influenced by the procedural regularity of the underlying domain rather than being uniformly beneficial across all tasks.

% 分析skill 节点 拓扑结构 和任务难度的正相关性
% \paragraph{Finding 8: skill topology to task hard analysis}
\paragraph{Task Topology Determines Skill Effectiveness.}
Task topology exhibits a clear correlation with skill effectiveness (Figure~\ref{fig:analysis}(b)). 
Single skill tasks receive almost no benefit, indicating that isolated procedures can already be handled reliably by modern foundation models. 
As dependency structures become more compositional, performance improvements increase substantially. Chain, Fan-out, and especially Fan-in topologies consistently obtain the largest gains across most models, suggesting that skills are particularly effective at encoding reusable coordination patterns across multiple dependent subtasks. Diamond tasks, however, display the greatest variance, with strong models benefiting considerably while weaker models occasionally suffering performance degradation. This observation indicates that executing highly compositional skills requires not only well-structured procedural knowledge but also sufficient instruction-following and coordination capabilities from the underlying model. Overall, skills can provide their greatest value when tasks involve complex compositions of reusable operations rather than isolated actions.

\begin{figure*}[t]
    \centering
    
    \includegraphics[width=1\textwidth]{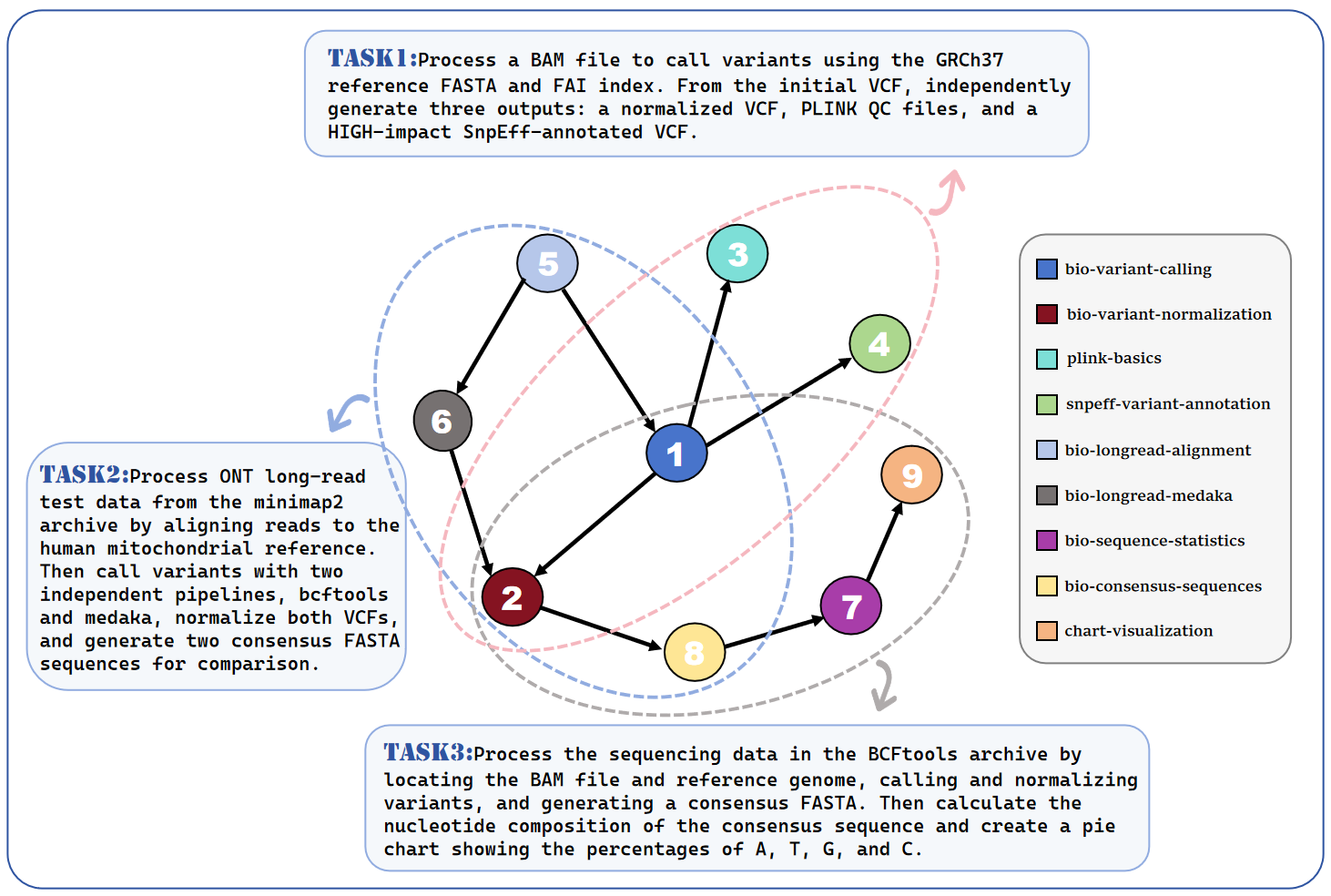}
    \caption{Case Study: From SkillNet subgraphs to synthesized tasks.}
    \label{fig:case_study}
\end{figure*}

\paragraph{Case Study: From SkillNet to Synthesized Tasks}
We provide an example of the conversion from SkillNet to task instructions, as shown in Figure~\ref{fig:case_study}.

\textbf{Furthermore, \dataname can continuously synthesize new benchmarks and evolve together with the skill ecosystem.}

\section{SkillNet-Fabric}
\label{sec:skillnet-fabric}

\begin{table}[t]
    \centering
    \makebox[\textwidth][c]{%
        \begin{minipage}[b]{0.51\textwidth}
            \centering
            \includegraphics[width=\linewidth]{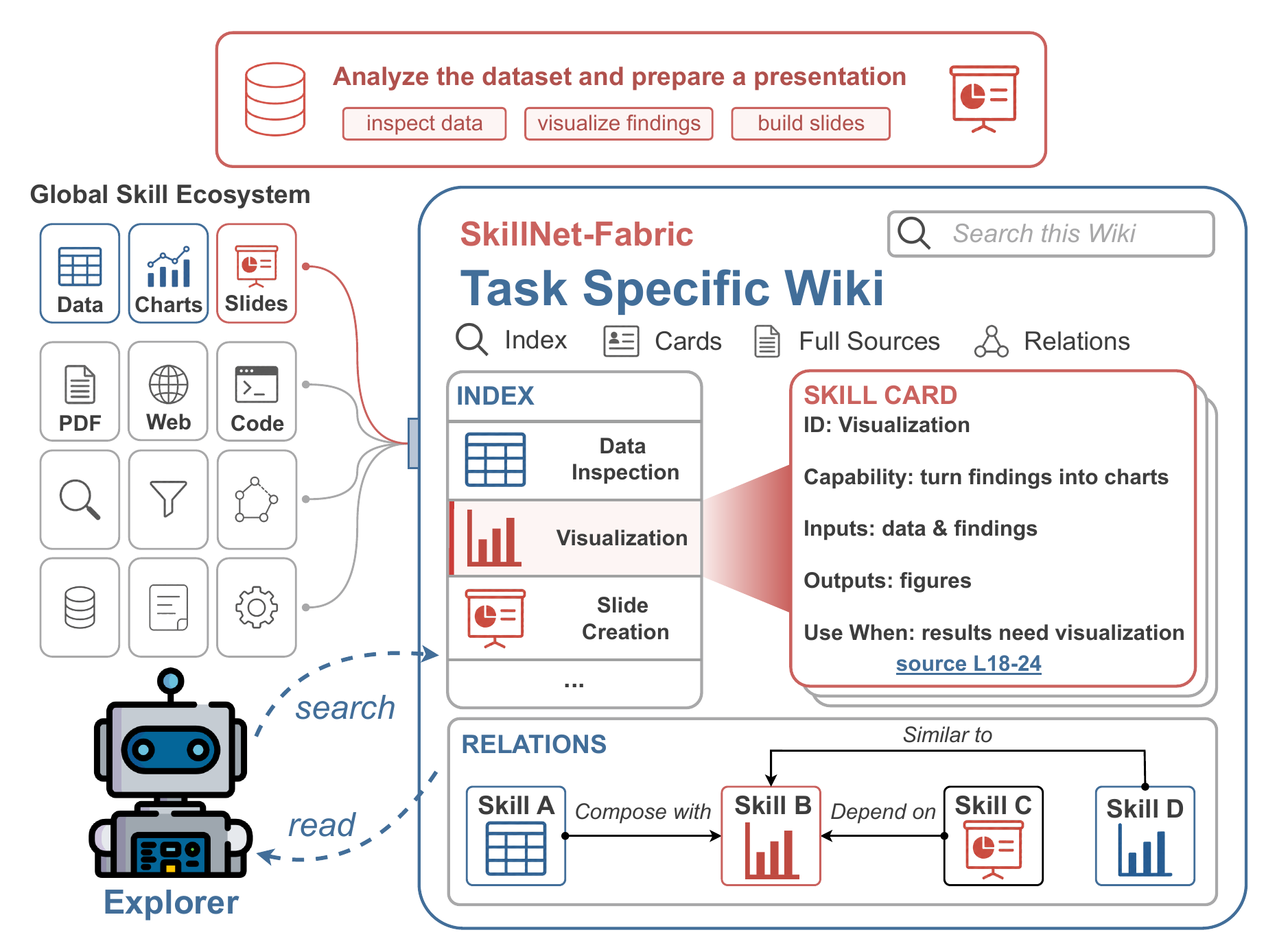}
        \end{minipage}%
        \hspace{0.02\textwidth}%
        \begin{minipage}[b]{0.47\textwidth}
            \centering
            \small
            \renewcommand{\arraystretch}{1.05}
            \setlength{\tabcolsep}{2pt}
            \begin{tabular*}{\linewidth}[b]{@{\extracolsep{\fill}}lcc@{}}
                \toprule
                \textbf{Methods} & \textbf{Recall} $\uparrow$ & \textbf{FullCoverage} $\uparrow$ \\
                \midrule
                \multicolumn{3}{c}{\textbf{$K=5$}} \\
                \midrule
                BM25 & 30.00 & 20.00 \\
                BGE Large v1.5 & 54.40 & 37.33 \\
                Qwen3 E$\times$R & 61.40 & 42.67 \\
                SkillRouter & 63.40 & 46.67 \\
                \textbf{SkillNet-Fabric} & \textbf{67.91} & \textbf{49.33} \\
                \midrule
                \multicolumn{3}{c}{\textbf{$K=10$}} \\
                \midrule
                BM25 & 36.50 & 24.00 \\
                BGE Large v1.5 & 61.10 & 41.33 \\
                Qwen3 E$\times$R & 67.77 & 49.33 \\
                SkillRouter & 69.42 & 50.67 \\
                \textbf{SkillNet-Fabric} & \textbf{74.80} & \textbf{60.00} \\
                \bottomrule
            \end{tabular*}
        \end{minipage}%
    }

    \makebox[\textwidth][c]{%
        \begin{minipage}[t]{0.51\textwidth}
            \captionof{figure}{Framework of SkillNet-Fabric. A task specific Wiki provides the Explorer with retained candidates and evidence for final skill selection.}
            \label{fig:skillnet-fabric-wiki-overview}
        \end{minipage}%
        \hspace{0.02\textwidth}%
        \begin{minipage}[t]{0.47\textwidth}
            \captionof{table}{Skill routing on SkillRouter Hard. Recall and FullCoverage are macro percentages over 75 tasks. Qwen3 E$\times$R denotes Qwen3 Embedding $\times$ Qwen3 Ranker.}
            \label{tab:skillnet-fabric-routing}
        \end{minipage}%
    }
\end{table}

\subsection{Overview}

SkillNet-Gym shows that agents can benefit from curated skills, but identifying a compact skill set whose capabilities jointly support a task remains difficult in large community skill pools. We therefore introduce \textbf{SkillNet-Fabric}, a task time routing layer that places a task specific Wiki between the global skill ecosystem and final set formation. The Wiki brings together task relevant candidates and their supporting evidence, which the Explorer uses to form the final skill set. Figure~\ref{fig:skillnet-fabric-wiki-overview} summarizes this process from the global Wiki to the task specific routing context.

\subsection{Routing Task and Routing Space}
\label{sec:fabric-routing-space}

Let $\mathcal{S}=\{s_i\}_{i=1}^{N}$ denote the global SkillNet ecosystem. Given a task $q$ and selection budget $K$, a router $\pi$ returns a final set $\mathcal{Y}_q$. The objective is joint task support, meaning that the capabilities and instructions in $\mathcal{Y}_q$ should collectively cover the requirements expressed by $q$.

Before selecting $\mathcal{Y}_q$, the router determines a candidate set $\mathcal{S}_q$ and evidence $\mathcal{E}_q$ available for evaluating it. Together, these components define the routing space $\mathcal{R}_q$. The final set is then formed from this space:
\begin{equation}
    \begin{aligned}
        \mathcal{R}_q
        &= (\mathcal{S}_q,\mathcal{E}_q),
        \qquad \mathcal{S}_q\subseteq\mathcal{S},\\
        \mathcal{Y}_q
        &= \pi(q,\mathcal{R}_q;K),
        \qquad \mathcal{Y}_q\subseteq\mathcal{S}_q,
        \quad |\mathcal{Y}_q|\leq K.
    \end{aligned}
    \label{eq:fabric-routing-space}
\end{equation}
Recovering the required set therefore has two aspects: candidate availability, which determines the contents of $\mathcal{S}_q$, and final set formation, through which the Explorer constructs $\mathcal{Y}_q$ from those candidates. Expanding $\mathcal{S}_q$ can make more required skills available, but it can also introduce irrelevant candidates and context, making final set formation less focused~\citep{wang2026skilldocumentqueryconditionedcompatibility}.

\begin{figure}[ht]
    \centering
    \includegraphics[width=\textwidth]{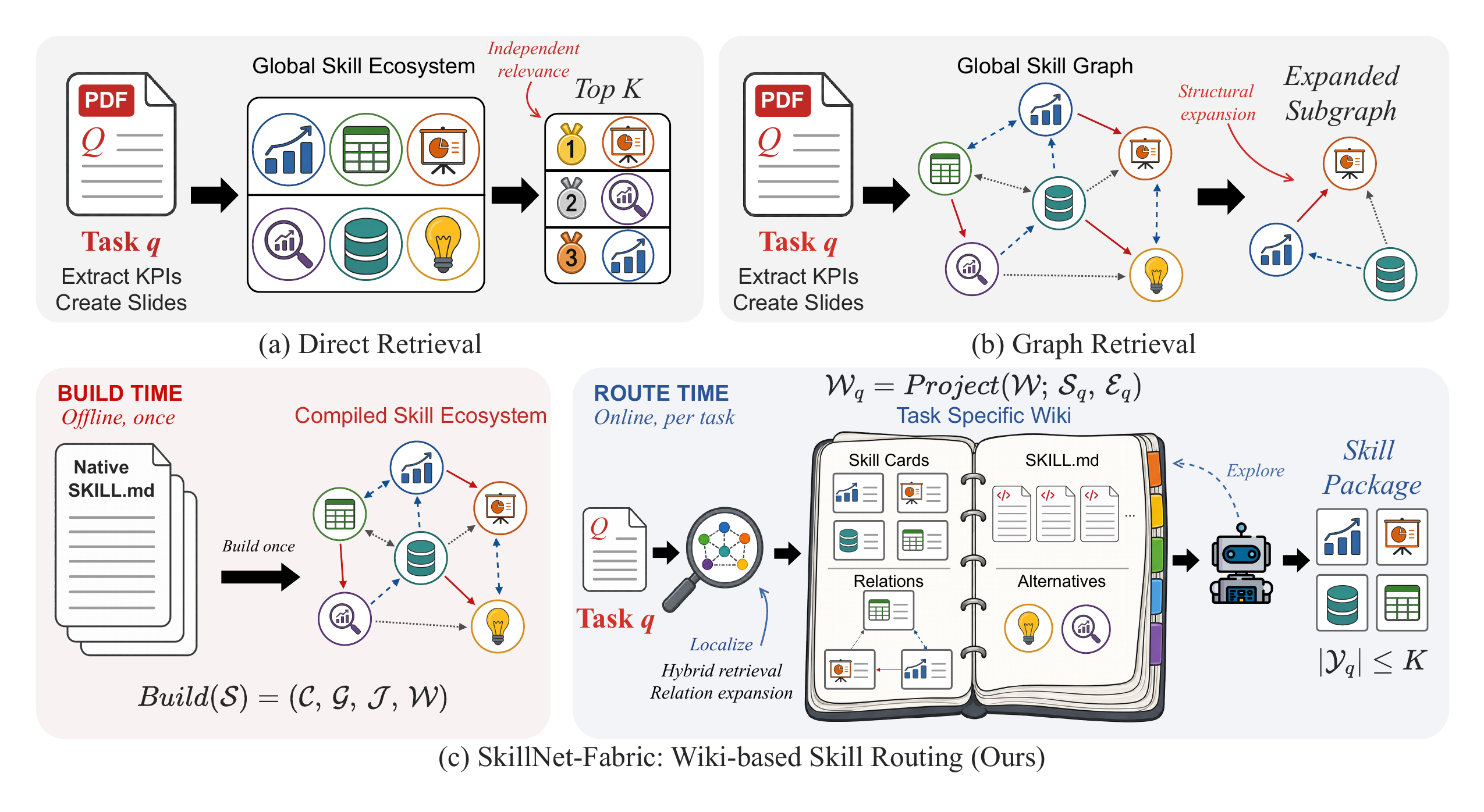}
    \caption{Comparison of two routing paradigms with SkillNet-Fabric.}
    \label{fig:skillnet-fabric-routing-overview}
\end{figure}

Routing methods construct the routing space in different ways. Figure~\ref{fig:skillnet-fabric-routing-overview} compares two routing paradigms with SkillNet-Fabric. Direct retrieval ranks and reranks candidates by task relevance, while graph retrieval also uses relations to expand or organize them~\citep{zheng2026skillrouterskillroutingllm,liu2026graphofskills}. SkillNet-Fabric combines retrieval and relation

\noindent expansion to construct $\mathcal{S}_q$, then materializes the retained candidates and evidence as a task specific Wiki. The Explorer navigates this Wiki to form the final set $\mathcal{Y}_q$.

\subsection{Build Time Skill Representation}
\label{sec:fabric-build}

During the build stage, heterogeneous skill sources are converted into reusable, source grounded representations. This stage produces contracts $\mathcal{C}$, a typed relation graph $\mathcal{G}$, retrieval indexes $\mathcal{J}$, and a global Wiki $\mathcal{W}$:
\begin{equation}
    (\mathcal{C},\mathcal{G},\mathcal{J},\mathcal{W})=\operatorname{Build}(\mathcal{S}).
    \label{eq:fabric-build}
\end{equation}

\paragraph{Skill Contracts.}
For each skill, its stated capability, use conditions, inputs, and outputs are extracted into a source grounded contract. The contract provides a compact representation for retrieval and relation construction, while its fields remain linked to supporting source lines.

\paragraph{Typed Relations.}
Relation construction first identifies likely related skill pairs from explicit references, compatible inputs and outputs, and lexical or semantic proximity. A language model then reads the two skill sources and classifies each pair as \texttt{depend\_on}, \texttt{compose\_with}, or \texttt{similar\_to}. Dependency and composition relations preserve their source and target roles, whereas similarity is symmetric and identifies possible alternatives. The resulting typed graph organizes the Wiki and supports candidate discovery and comparison.

\paragraph{Global Wiki.}
The global Wiki organizes the skill ecosystem into a navigable knowledge space grounded in the underlying skill sources. Its linked pages represent each skill with a card that summarizes its contract and links to the complete source, while an index connects these pages for navigation. Together, these pages provide the structured context that routing uses to project the task specific Wiki for the Explorer.

\subsection{Task Time Wiki Localization}
\label{sec:fabric-localization}

At task time, routing uses the retrieval indexes and typed relation graph to identify the candidate set and evidence available for a task. Relevant seeds are retrieved first and then expanded through graph relations. The retained candidates and evidence are projected from the global Wiki to form a task specific Wiki $\mathcal{W}_q$:
\begin{equation}
    \begin{aligned}
        (\mathcal{S}_q,\mathcal{E}_q)
        &=\operatorname{Localize}(q;\mathcal{G},\mathcal{J},L),\\
        \mathcal{W}_q
        &=\operatorname{Project}(\mathcal{W};\mathcal{S}_q,\mathcal{E}_q).
    \end{aligned}
    \label{eq:fabric-localize}
\end{equation}
The parameter $L$ sets the maximum number of candidates admitted to $\mathcal{W}_q$.

\paragraph{Hybrid Seed Retrieval.}
BM25 captures exact procedural terms~\citep{10.1561/1500000019}, whereas the dense index captures semantic matches. Reciprocal rank fusion combines the two rankings~\citep{10.1145/1571941.1572114}, and the fused ranking determines the seed set for relation expansion.

\paragraph{Relation Expansion.}
Starting from the seed set, relation expansion follows typed relations up to a fixed graph depth and admits connected candidates until $L$ candidates have been retained. This step supplements task relevance with related skills that may not appear among the initial retrieval results. Seeds are retained first, and expanded candidates fill the remaining capacity.

\paragraph{Wiki Projection.}
The task specific Wiki projects the retained candidates from the global Wiki and augments them with the retrieval and relation evidence used to admit them. Local relations support comparison among the retained skills.

\subsection{Wiki Exploration and Route Formation}
\label{sec:fabric-exploration}

\paragraph{Wiki Exploration.}
The Explorer turns the task specific Wiki into a final skill set. It starts from the Wiki index and contract cards, selectively reads complete sources, and consults local relations when candidates require comparison. Based on this evidence, it selects up to $K$ skills and records how each selected skill contributes to the task. The selected set is
\begin{equation}
    \mathcal{Y}_q=\operatorname{Explore}(q,\mathcal{W}_q;K).
    \label{eq:fabric-explore}
\end{equation}

\paragraph{Route Formation.}
The Planner combines the original task with the selected skills and their supporting evidence to produce an execution prompt for the downstream agent.

\subsection{Experimental Settings}
\label{sec:fabric-experimental-settings}

\paragraph{Benchmarks and Metrics.}
We evaluate SkillNet-Fabric at two levels: routing quality on SkillRouter Hard~\citep{zheng2026skillrouterskillroutingllm} and downstream task performance on SkillsBench~\citep{skillsbench2025} and AgentSkillOS~\citep{li2026organizingorchestratingbenchmarkingagent}. SkillRouter Hard contains 75 scored tasks over 79{,}141 candidate skills. We report macro Recall of required skills and FullCoverage, the fraction of tasks for which the complete required set is selected. SkillsBench contains 87 tasks with deterministic verifiers; its metric is the task macro mean verifier reward. AgentSkillOS contains 30 tasks across five categories and reports task level Bradley--Terry scores~\citep{19ff28b9-64f9-3656-ba40-08326a05748e} for skill pools of 53, 500, and 1{,}000 skills.

\paragraph{Baselines.}
For SkillRouter Hard, the comparison includes BM25~\citep{10.1561/1500000019}, BGE Large v1.5~\citep{10.1145/3626772.3657878}, Qwen3 Embedding with Qwen3 Ranker~\citep{zhang2025qwen3embeddingadvancingtext}, and the official SkillRouter method~\citep{zheng2026skillrouterskillroutingllm}. SkillsBench compares SkillNet-Fabric with execution without skills, Golden Skills, SkillRouter, and SkillsVote~\citep{liu2026skillsvotelifecyclegovernanceagent}; Golden Skills denotes the reference skill set. AgentSkillOS compares SkillNet-Fabric with execution without skills, exposure to the full skill pool, and the native AgentSkillOS system.

\paragraph{Configuration.}
For task time routing, we use BGE M3 embeddings~\citep{chen-etal-2024-m3} and expand relations to depth two. On SkillRouter Hard, GPT-5.6 Terra~\citep{openai2026gpt56} serves as the Explorer with 100 retrieval seeds, $L=1{,}000$, and $K\in\{5,10\}$. SkillsBench uses the same localization configuration with $K=10$. For downstream evaluation, we use GPT-5.4 mini~\citep{openai2026gpt54MiniNano} and GPT-5.6 Terra, with the same model performing routing, planning, and execution within each condition. AgentSkillOS uses eight seeds, $L=100$, and $K=8$.

\subsection{Main Results}
\label{sec:fabric-main-results}

\paragraph{Routing Quality on SkillRouter Hard.}
SkillNet-Fabric achieves the highest required skill recall and complete set recovery among the compared methods at both selection budgets. Table~\ref{tab:skillnet-fabric-routing} reports macro Recall and FullCoverage when the router selects from 79{,}141 candidate skills. At $K=5$, SkillNet-Fabric reaches 67.91\% Recall and 49.33\% FullCoverage, exceeding the official SkillRouter method by 4.51 and 2.66 percentage points, respectively. Increasing the budget to $K=10$ raises the scores to 74.80\% and 60.00\%, widening the margins to 5.38 and 9.33 points. The wider FullCoverage margin suggests that additional selection capacity is particularly useful for recovering complementary skills needed for complete set formation.

\paragraph{Downstream Task Performance.}
We next evaluate whether the selected routes improve downstream performance. Table~\ref{tab:skillnet-fabric-downstream} reports results on SkillsBench and AgentSkillOS.
\begin{table}[!t]
    \centering
    \small
    \renewcommand{\arraystretch}{1.05}
    \setlength{\tabcolsep}{4pt}
    \begin{tabular*}{\textwidth}{@{\extracolsep{\fill}}lcccc@{}}
        \toprule
        \rowcolor{mydarkbrown!40}
        \multicolumn{5}{c}{\textbf{(a) SkillsBench: Mean Verifier Reward (\%)}} \\
        \midrule
        \textbf{Methods} & \textbf{Easy} $\uparrow$ & \textbf{Medium} $\uparrow$ & \textbf{Hard} $\uparrow$ & \textbf{Overall} $\uparrow$ [95\% CI] \\
        \midrule
        \multicolumn{5}{c}{\textbf{GPT-5.4 mini}} \\
        \midrule
        w/o Skills & 47.22 & 27.73 & 29.46 & 29.63 [22.54, 37.09] \\
        Golden Skills & 48.14 & \textbf{50.13} & 32.20 & 44.22 [35.89, 52.51] \\
        SkillRouter & 42.59 & 42.18 & \textbf{40.54} & 41.68 [33.26, 50.21] \\
        SkillsVote & 40.74 & 34.91 & 33.93 & 34.99 [27.30, 42.93] \\
        \textbf{SkillNet-Fabric} & \textbf{62.96} & 48.45 & 37.98 & \textbf{46.08} [37.51, 54.62] \\
        \midrule
        \multicolumn{5}{c}{\textbf{GPT-5.6 Terra}} \\
        \midrule
        w/o Skills & 52.77 & 46.06 & 45.24 & 46.26 [37.12, 55.53] \\
        Golden Skills & 48.14 & 66.17 & 52.92 & 60.66 [51.90, 69.29] \\
        SkillRouter & 53.70 & 59.66 & 46.49 & 55.01 [45.82, 63.95] \\
        SkillsVote & 59.26 & 60.34 & 50.42 & 57.07 [47.89, 65.89] \\
        \textbf{SkillNet-Fabric} & \textbf{69.44} & \textbf{67.92} & \textbf{63.69} & \textbf{66.67} [58.75, 74.58] \\
        \midrule
        \rowcolor{sky!40}
        \multicolumn{5}{c}{\textbf{(b) AgentSkillOS: Bradley--Terry Score}} \\
        \midrule
        \textbf{Methods} & \textbf{Seed Pool} $\uparrow$ & \textbf{500 Skills} $\uparrow$ & \textbf{1,000 Skills} $\uparrow$ & \textbf{Pool Mean} $\uparrow$ \\
        \midrule
        \multicolumn{5}{c}{\textbf{GPT-5.4 mini}} \\
        \midrule
        w/o Skills & 33.16 & 42.16 & 46.96 & 40.76 \\
        Full Pool & 53.85 & 50.99 & 46.30 & 50.38 \\
        AgentSkillOS & 61.11 & 46.96 & 53.13 & 53.73 \\
        \textbf{SkillNet-Fabric} & \textbf{61.89} & \textbf{65.20} & \textbf{65.54} & \textbf{64.21} \\
        \midrule
        \multicolumn{5}{c}{\textbf{GPT-5.6 Terra}} \\
        \midrule
        w/o Skills & 34.42 & 30.03 & 39.34 & 34.60 \\
        Full Pool & 28.38 & 46.29 & 34.28 & 36.32 \\
        AgentSkillOS & 40.84 & 44.95 & 40.43 & 42.07 \\
        \textbf{SkillNet-Fabric} & \textbf{90.03} & \textbf{78.53} & \textbf{80.45} & \textbf{83.00} \\
        \bottomrule
    \end{tabular*}
    \caption{Downstream task performance on SkillsBench and AgentSkillOS.
    Panel (a) reports task macro mean verifier rewards over 87 SkillsBench
    tasks; brackets contain 95\% confidence intervals for the Overall score.
    Panel (b) reports task level Bradley--Terry scores rescaled to 0 to 100 and
    averaged over 30 AgentSkillOS tasks across pools of 53, 500, and 1{,}000
    skills.}
    \label{tab:skillnet-fabric-downstream}
\end{table}
On SkillsBench, SkillNet-Fabric achieves the highest overall verifier reward in both model conditions: 46.08\% with GPT-5.4 mini and 66.67\% with GPT-5.6 Terra. It exceeds SkillRouter by 4.40 percentage points in the former condition and SkillsVote by 9.60 points in the latter, the strongest routing baselines in their respective settings. On AgentSkillOS, SkillNet-Fabric ranks first in all six combinations of model and skill pool, with Pool Mean scores of 64.21 for GPT-5.4 mini and 83.00 for GPT-5.6 Terra. Under GPT-5.4 mini, its margin over the strongest comparator grows from 0.78 points on the seed pool to 14.21 and 12.41 points on the pools of 500 and 1{,}000 skills.

\begin{table}[H]
    \centering
    \small
    \renewcommand{\arraystretch}{1.12}
    \setlength{\tabcolsep}{4pt}
    \begin{tabular*}{\textwidth}{@{\extracolsep{\fill}}lcccc
        >{\centering\arraybackslash}m{1.55cm}
        >{\centering\arraybackslash}m{1.55cm}c@{}}
        \toprule[1.5pt]
        \multirow{2}{*}{\textbf{Variant}}
        & \multicolumn{4}{c}{\textbf{Wiki Components}}
        & \multicolumn{2}{c}{\textbf{Downstream Quality}}
        & \multicolumn{1}{c}{\textbf{Exploration}} \\
        \cmidrule(lr){2-5} \cmidrule(lr){6-7} \cmidrule(lr){8-8}
        & \makecell{Graph\\Expansion}
        & \makecell{Relation\\Evidence}
        & \makecell{Contract\\Cards}
        & \makecell{Complete\\Sources}
        & \makecell[c]{BT Score}
        & \makecell[c]{$\Delta$}
        & \makecell[c]{Tokens (K)} \\
        \midrule
        w/o Graph Expansion & \xmark & \cmark & \cmark & \cmark & 54.15 & $-11.36$ & 16.66 \\
        w/o Relation Evidence & \cmark & \xmark & \cmark & \cmark & 53.05 & $-12.46$ & 30.26 \\
        w/o Contract Cards & \cmark & \cmark & \xmark & \cmark & 49.80 & $-15.71$ & 34.23 \\
        w/o Complete Sources & \cmark & \cmark & \cmark & \xmark & 48.12 & $-17.39$ & 24.42 \\
        \midrule[0.3pt]
        \textbf{Full Wiki} & \cmark & \cmark & \cmark & \cmark & \textbf{65.51} & 0.00 & 27.82 \\
        \bottomrule[1.5pt]
    \end{tabular*}
    \caption{Wiki component ablation on AgentSkillOS. ``{\cmark}'' and ``{\xmark}''
    indicate whether each component is present or removed. $\Delta$ is relative
    to the full Wiki. Explorer tokens are averaged per task and reported in
    thousands.}
    \label{tab:skillnet-fabric-ablation}
\end{table}
\begin{figure}[H]
    \centering
    \includegraphics[width=\textwidth]{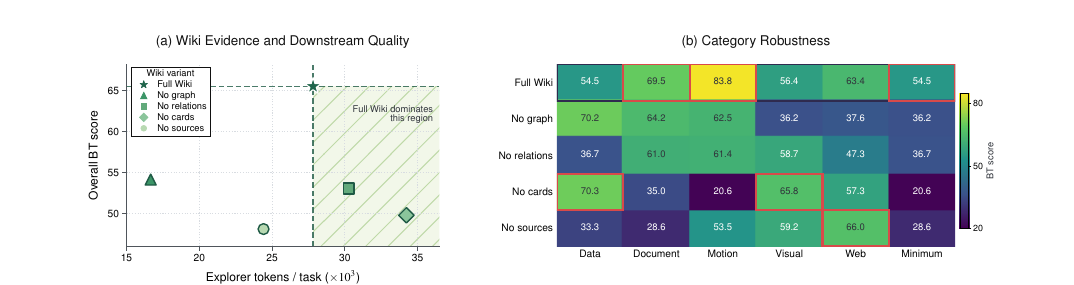}
    \caption{Effects of Wiki components on AgentSkillOS. Panel (a) relates
    overall Bradley--Terry score to Explorer tokens per task. Panel (b) reports
    category scores and the minimum across categories.}
    \label{fig:skillnet-fabric-wiki-analysis}
\end{figure}
\FloatBarrier
\subsection{Analysis}
\label{sec:fabric-analysis}

\paragraph{Wiki Components.}
To isolate the contribution of individual Wiki components to downstream task performance, we compare the complete Wiki with variants that remove graph expansion, relation evidence, contract cards, or complete sources. The evaluation covers all 30 AgentSkillOS tasks with GPT-5.4 mini and the pool of 500 skills; Table~\ref{tab:skillnet-fabric-ablation} reports the overall Bradley--Terry score and mean Explorer tokens for each variant.

The complete Wiki obtains a Bradley--Terry score of 65.51, at least 11.36 points above every ablation. The ablations reveal that downstream quality does not follow Explorer token use alone. Removing contract cards or relation evidence increases token use to 34.2K and 30.3K per task, respectively, while scores fall to 49.80 and 53.05. Removing graph expansion or complete sources reduces token use to 16.7K and 24.4K, but scores also fall to 54.15 and 48.12. The relationship between quality and Explorer tokens in Figure~\ref{fig:skillnet-fabric-wiki-analysis}(a) therefore shows that the Wiki's benefit depends on the composition of the available context. Graph expansion and complete sources affect which candidates can be inspected and how they can be assessed, while cards and relation evidence provide compact cues for navigating and comparing them.

The category results in Figure~\ref{fig:skillnet-fabric-wiki-analysis}(b) show that the components contribute differently across task categories. The variant without graph expansion remains strong on Data (70.2) but is weak on Visual (36.2) and Web (37.6). The variant without contract cards produces a similar contrast, with 70.3 on Data and 65.8 on Visual but only 20.6 on Motion. The variant without complete sources leads on Web (66.0) while falling to 33.3 on Data and 28.6 on Document. In contrast, the complete Wiki maintains a minimum category score of 54.45, compared with 36.73 for the strongest minimum among the ablations. No individual component helps every task type uniformly; the complete Wiki gives the most stable coverage across categories.

\paragraph{Candidate Availability and Final Set Formation.}
We further study how the localization limit affects candidate availability and final set formation on the 75 SkillRouter Hard tasks. We vary $L\in\{10,100,1{,}000,5{,}000,10{,}000\}$ and compare these settings with the full pool. For each task and value of $L$, GPT-5.6 Luna, GPT-5.6 Terra, and GPT-5.6 Sol receive the same task specific Wiki and select at most ten skills. Candidate FullCoverage measures whether every required skill is available before exploration, whereas Final FullCoverage@10 measures whether
\begin{figure}[H]
    \centering
    \includegraphics[width=\textwidth]{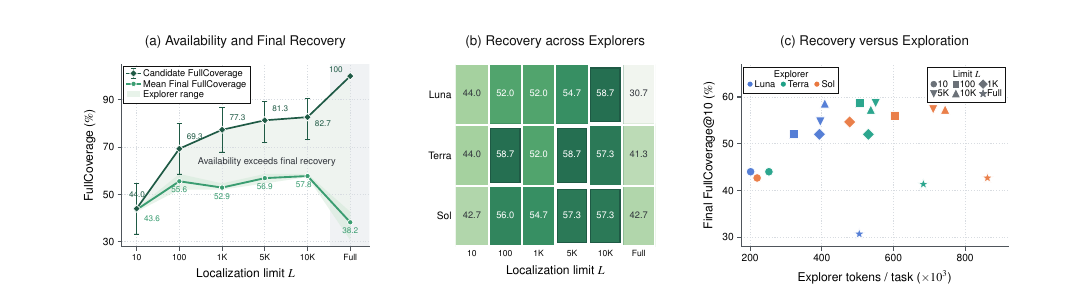}
    \caption{Candidate availability and final set formation on SkillRouter
    Hard. Panel (a) compares Candidate FullCoverage with mean Final
    FullCoverage@10 as $L$ grows. Panel (b) reports Final FullCoverage@10 by
    Explorer, and panel (c) relates Final FullCoverage@10 to Explorer tokens. Full
    denotes all 79{,}141 skills.}
    \label{fig:skillnet-fabric-routing-analysis}
\end{figure}
\noindent the Explorer forms the complete required set. Figure~\ref{fig:skillnet-fabric-routing-analysis} summarizes the resulting relationship between candidate availability, final set formation, and exploration effort.
The sweep separates candidate availability from final set formation. Candidate FullCoverage rises from 44.00\% at $L=10$ to 69.33\% at $L=100$, 81.33\% at $L=5{,}000$, and 82.67\% at $L=10{,}000$, with only a 1.33 point increase in the last interval. The full pool makes all required skills available and reaches 100\% Candidate FullCoverage. Final FullCoverage@10 follows a different trajectory: the mean increases from 43.56\% at $L=10$ to 57.78\% at $L=10{,}000$, then drops to 38.22\% with the full pool. At $L=10$, the two measures are nearly identical, so most failures reflect limited candidate availability. Under the full pool, their 61.78 point gap shows that the remaining failures arise during final set formation, when the Explorer compares candidates. Figure~\ref{fig:skillnet-fabric-routing-analysis}(a) visualizes this separation.

Results across Explorers show that the preferred value of $L$ varies by model. Luna improves as the candidate space grows and reaches 58.67\% at $L=10{,}000$; Terra reaches 58.67\% at both $L=100$ and $L=5{,}000$; and Sol peaks at 57.33\% at both $L=5{,}000$ and $L=10{,}000$. For each Explorer, the best tested $L$ setting outperforms the corresponding full pool condition. Figure~\ref{fig:skillnet-fabric-routing-analysis}(b) shows these differences across Explorers. With the routing procedure and selection budget held fixed, the different peaks suggest that the preferred scope depends on the Explorer model's capacity to compare the available candidates.

Exploration effort follows the same pattern. At each Explorer's best tested $L$ setting, token use is lower and Final FullCoverage@10 is higher than under the full pool, as shown in Figure~\ref{fig:skillnet-fabric-routing-analysis}(c). A larger context can therefore increase Explorer token use without improving complete set recovery. Taken together, the results support the two stage routing design. Increasing $L$ improves candidate availability, while an excessively broad space adds candidates and context that can make final set formation less focused. The task specific Wiki connects these stages by presenting the Explorer with the candidates and evidence retained during routing, and its most effective scope depends on the Explorer model in the tested settings.

\section{Application Scenarios}
% 总体介绍下 我们这个支持很多场景 下面重点介绍science和coding这俩

\begin{figure}[ht]
    \centering
    \includegraphics[width=0.95\textwidth]{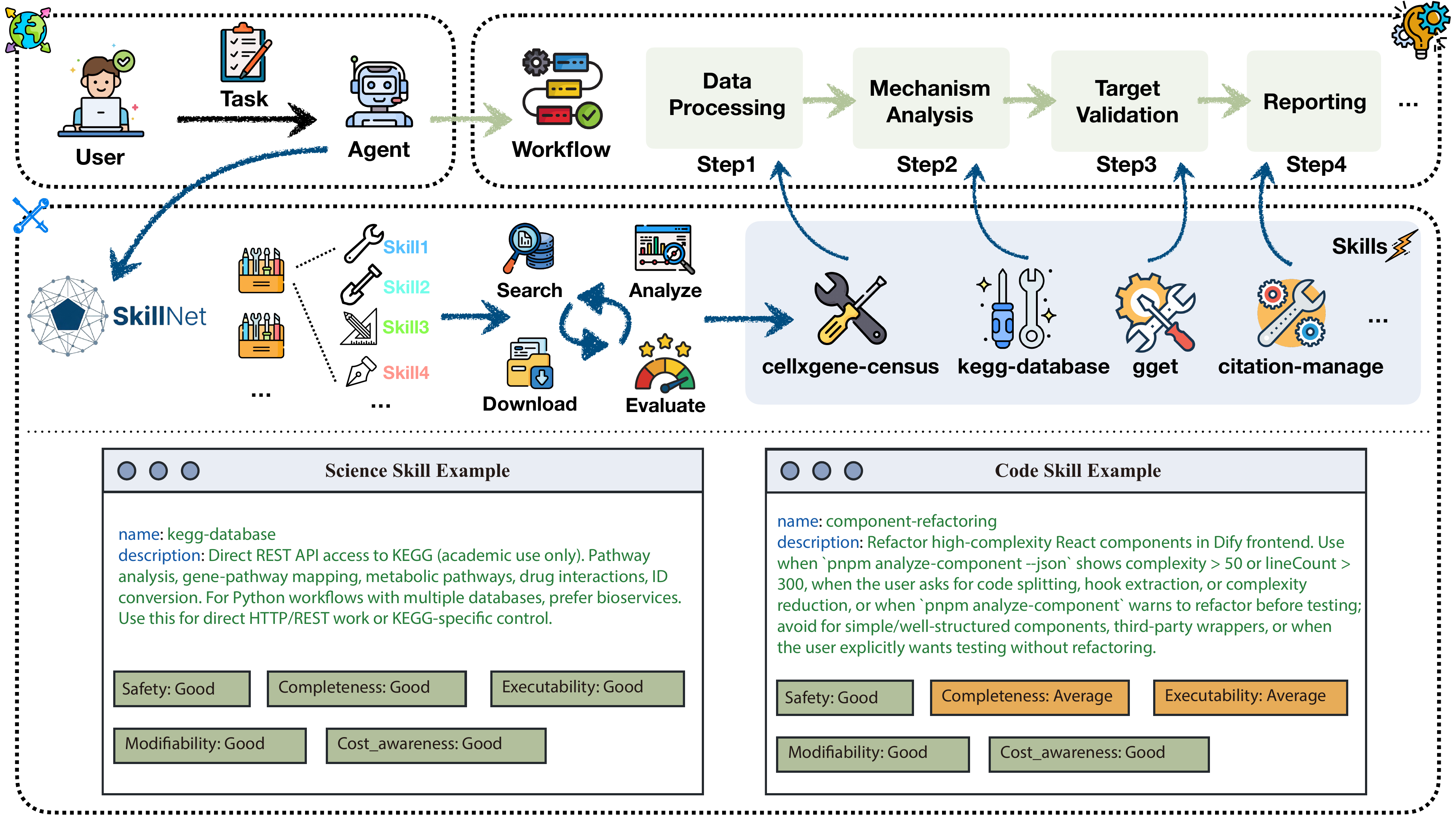} 
    \caption{Example of SkillNet application scenarios. The framework decomposes user task into actionable steps (top), with representative skill acquisition and multi-dimensional evaluations for Science and Coding scenarios (bottom).}
    \label{fig:skillnet_scenario}
\end{figure}

% SkillNet enables the construction of complex workflows for a wide range of applications. We detail below its specific utility in Science and Coding scenarios.
As illustrated in Figure~\ref{fig:skillnet_scenario}, SkillNet bridges the gap between high-level user intentions and executable agent actions by organizing specialized skills into a coherent workflow. 
We detail below its specific utility in autonomous scientific discovery and coding agent scenarios. 
Note that the examples are illustrative prototypes and do not represent real-world applications.

\subsection{SkillNet for Autonomous Scientific Discovery}

In this section, we demonstrate how SkillNet organizes heterogeneous agent skills into a coherent, executable, and evaluable research workflow. 
This system performs the scientific task of identifying potential disease-associated genes and candidate therapeutic targets from large-scale biological data, ultimately generating research reports characterized by scholarly readability.

SkillNet first schedules data processing skills to perform cleaning and clustering on single-cell RNA-seq data, thereby identifying key genetic markers. 
Subsequently, SkillNet invokes mechanistic analysis and target validation skills to map these genes onto biological pathways and cross-validate their clinical significance. Finally, SkillNet activates report generation skills to integrate fragmented analytical results into structured scientific documents with formal citations.

This process profoundly reflects the core value of SkillNet: transforming fragmented research skills into a structured, composable network of capabilities. This architecture enables an AI Scientist to transcend the limitations of single-domain tasks and engage meaningfully in the complete closed-loop process of complex scientific discovery.

\subsection{SkillNet for Autonomous Coding Agent}

% SkillNet 驱动的编码智能体：面向复杂软件工程任务的能力组织范式

% 为进一步验证 SkillNet 的通用性，我们构建了一个面向大规模软件工程任务的编码场景，展示 SkillNet 在复杂代码理解、重构与功能开发中的应用潜力。该场景的目标是在既有生产级代码系统中完成特性扩展与结构演化，同时保证正确性、性能与长期可维护性。

% 工作流从代码库的结构化理解开始。SkillNet 调度代码分析技能，对源代码进行解析，构建抽象语法树与模块依赖图，并自动归纳系统的架构模式与设计假设。这一阶段的核心在于将非结构化的代码文本转化为机器可理解的结构化表示，为后续推理与决策提供基础。

% 在此基础上，SkillNet 进一步协调需求分解与影响分析相关的推理技能。系统将高层功能需求映射为具体的代码级修改目标，并通过依赖分析识别受影响模块、潜在回归风险以及必要的接口变更。这一过程避免了单一智能体对复杂工程任务的过载依赖，而是将理解、规划与风险评估分布到多个专用技能之中。

% 在实现阶段，SkillNet 将代码生成、测试与验证技能组织为闭环的迭代开发流程。生成的代码修改会同步触发静态分析、单元测试生成与回归测试执行，测试结果与执行轨迹被持续反馈至技能图中，用于驱动自动修正或替代实现路径的选择。

% 最终，SkillNet 调用面向文档与维护的技能，自动生成重构说明、接口变更记录以及架构更新摘要。所有代码变更均与其需求来源和执行历史建立可追溯关联，为系统的长期演化与协作开发提供支撑。

% 该编码场景表明，SkillNet 将软件工程问题重新建模为能力组合与协同的问题，而非简单的提示驱动代码生成。通过将编码、分析、测试与文档等能力组织为结构化的 Agent Skills，SkillNet 为复杂工程任务提供了一种可靠、可扩展的自动化范式。 

% 在本小节中，我们构建了一个面向大规模软件工程任务的编码场景，展示 SkillNet 在复杂代码理解、重构与功能开发中的应用潜力。该场景的目标是在既有生产级代码系统中完成特性扩展与结构演化，同时保证正确性、性能与长期可维护性。
% SkillNet 首先调度代码分析技能，构建系统架构的结构化表示；随后协调需求分解与影响分析技能，将高层需求精准映射为代码级修改目标，并评估回归风险；在实现阶段，SkillNet 将生成、测试与验证技能组织为闭环迭代流程，并利用实时反馈驱动自动修正，最终调用维护技能生成具备可追溯性的架构更新文档。
% 该场景证明了 SkillNet 的核心价值：将复杂的软件工程问题重新建模为能力组合与协同的问题。通过将编码、分析与测试等碎片化技能重构为结构化的能力网络，SkillNet 为复杂工程任务提供了一种可靠且可扩展的自动化范式，使 AI 能够真正胜任系统级的软件演进工作。

In this section, we construct a coding scenario tailored for large-scale software engineering tasks to demonstrate the potential of SkillNet in complex code comprehension, refactoring, and functional development. 
The objective of this scenario is to execute feature expansion and structural evolution within an existing production-grade codebase while ensuring correctness, performance, and long-term maintainability.

The workflow begins with SkillNet scheduling code analysis skills to construct a structured representation of the system architecture. Subsequently, it coordinates requirements decomposition and impact analysis skills to precisely map high-level functional requirements to specific code-level modification targets while assessing regression risks. During the implementation phase, SkillNet organizes generation, testing, and validation skills into a closed-loop iterative process, utilizing real-time feedback to drive automated corrections. Finally, it invokes maintenance skills to generate architectural update documentation characterized by complete traceability.

This scenario illustrates the core value of SkillNet: remodeling complex software engineering challenges as problems of capability composition and coordination. 
By restructuring fragmented skills such as coding, analysis, and testing into a structured network of agent skills, SkillNet provides a reliable and scalable automation paradigm for complex engineering tasks, enabling AI to competently perform system-level software evolution.

\subsection{Using OpenClaw with SkillNet}

% 在本小节中，我们展示了 SkillNet 如何与 OpenClaw 集成。OpenClaw 是一个高度可定制的开源个人 AI 代理框架。OpenClaw 采用技能延迟加载设计：在会话初始化时仅将紧凑的技能元数据注入系统提示词，完整指令在触发时按需加载。SkillNet 作为一项技能集成到该框架中，为代理扩展了动态技能获取、质量感知的技能库管理以及经验驱动的知识创建能力。
% 安装后，SkillNet 引入三种基于对话上下文自动激活的核心行为模式。第一，当用户提出复杂或陌生的任务时，代理会对 SkillNet 仓库执行任务前搜索以定位相关技能，随后将匹配的技能下载到本地工作区并读取对应的 SKILL.md 以提取可用的模式与工具配置。此外，当用户提供 GitHub 仓库链接或分享 PDF 等文档时，代理可直接调用 skillnet create 从该来源生成结构化技能，使外部知识立即可用于当前任务。第二，当用户请求技能库整理或质量审计时，代理调用 SkillNet 的 analyze 与 evaluate 命令生成结构化报告，包括多维度质量评分以及捕获 similar_to、belong_to、compose_with 和 depend_on 关系的技能间关系图谱。第三，当任务完成后其解决方案涉及非显而易见的可复用知识，或用户明确要求沉淀经验时，代理主动调用 skillnet create 将解决方案封装为标准化技能，并运行自动评估以在入库前验证其质量。
% 这一集成建立了一个闭环：社区贡献的技能指导任务执行，成功的任务成果被沉淀为新技能，定期分析则维护仓库质量。OpenClaw 与 SkillNet 的结合不再将每次对话视为孤立事件，而是通过结构化的技能流转将通用代理转变为持续自我进化的系统，直接支撑了可扩展、不断演进的代理智能愿景。

In this section, we demonstrate how SkillNet integrates with OpenClaw\footnote{\url{https://github.com/openclaw/openclaw}.}, a highly customizable open-source personal AI agent framework. 
OpenClaw adopts a lazy-loading skill design: at session initialization, only compact skill metadata is injected into the system prompt, and full instructions are loaded on demand when triggered. 
SkillNet is integrated as a skill in this framework, extending the agent with dynamic skill acquisition, quality-aware library management, and experience-driven knowledge creation.

Once installed, SkillNet introduces three core behavioral patterns that are activated automatically based on conversational context. 
First, when a user poses a complex or unfamiliar task, the agent performs a pre-task search against the SkillNet repository to locate relevant skills. 
It then downloads the matched skills into the local workspace and reads the retrieved SKILL.md to extract applicable patterns and tool configurations. 
In addition, when the user provides a GitHub repository URL or shares a document such as a PDF, the agent can directly invoke skillnet create to generate a structured skill from that source, making the external knowledge immediately available for the current task. 
Second, when the user requests skill library organization or quality auditing, the agent invokes SkillNet's analyze and evaluate commands to generate structured reports, which include multi-dimensional quality scores along with inter-skill relationship graphs that capture similar\_to, belong\_to, compose\_with, and depend\_on edges.
Third, after completing a task whose solution involved non-obvious or reusable knowledge, or when the user explicitly requests experience consolidation, the agent proactively invokes skillnet create to package the solution as a standardized skill and runs automatic evaluation to verify its quality before admitting it into the repository.

This integration establishes a closed loop where community-contributed skills inform task execution, successful outcomes are consolidated into new skills, and periodic analysis maintains repository quality. 
Rather than treating each conversation as an isolated episode, the OpenClaw + SkillNet combination transforms a general-purpose agent into a continuously self-improving system through structured skill circulation, directly supporting the vision of scalable, evolving agent intelligence.
\section{Related Work}

\subsection{Experience Consolidation and Skill Abstraction}

LLM-based agents have been extended with tools, planning, and memory to perform long-horizon tasks in complex environments \cite{xi2023risepotentiallargelanguage,park2023generativeagentsinteractivesimulacra,qin2023toolllmfacilitatinglargelanguage,wang2023voyageropenendedembodiedagent,zhu2023ghostminecraftgenerallycapable}. Several works have explored how agents can acquire and refine skills from interaction and experience. Reflexion \cite{shinn2023reflexionlanguageagentsverbal} and Expel \cite{zhao2024expelllmagentsexperiential} investigate how agents can summarize failures and extract corrective feedback. Memory-centric methods aim to accumulate long-term experience to support continual learning \cite{wang2024agentworkflowmemory,fang2026mempexploringagentprocedural,zhang2025memgenweavinggenerativelatent,ouyang2025reasoningbankscalingagentselfevolving}. 
Other studies explore structured skill learning and cross-task reuse, promoting generalization and transfer across domains \cite{chen2026cuaskilldevelopskillscomputer,wang2025reinforcementlearningselfimprovingagent,zheng2025skillweaverwebagentsselfimprove,yu2025polyskilllearninggeneralizableskills,tang2025agentkbleveragingcrossdomain,wu2026skillbasedautonomousagentsmaterial,xia2026skillrlevolvingagentsrecursive,zhang2026memskilllearningevolvingmemory}.
Despite these advances, most approaches still represent skills implicitly by encoding them in prompts, latent memories, or loosely organized workflows, thereby hindering systematic consolidation, evaluation, and reuse.

% \subsection{Skill Repositories and Evaluations}

% Parallel to the development of agentic systems, a number of community-driven repositories \cite{skillsmp2026,skillhub2026,skillssh2026} and evaluations \cite{skillsbench2025,liu2026agentskillswildempirical,ling2026agentskillsdatadrivenanalysis} have emerged to curate agent skills, aiming to promote reuse, standardization, and ecosystem-level collaboration. These platforms provide valuable resources for sharing prompts, tool wrappers, and task-solving templates. However, they largely rely on manual curation and ad hoc quality control, offering limited support for automated evaluation, safety verification, or long-term maintenance. 
% As a result, skill collections often suffer from redundancy, brittleness, and poor composability, restricting their scalability in large agent populations. Moreover, evaluation practices primarily focus on end-task performance, providing only indirect signals of skill quality, and fail to offer comprehensive insights into intrinsic properties such as safety, completeness, executability, modifiability, and cost-awareness. 
% These limitations motivate the need for a unified framework for skill organization, verification, and evolution.

\subsection{Skill Repositories and Evaluations}

\begin{table*}[htbp]
\centering
\resizebox{\textwidth}{!}{
\begin{tabular}{>{\raggedright\arraybackslash}p{3cm} >{\raggedright\arraybackslash}p{4cm} >{\raggedright\arraybackslash}p{2.5cm} >{\raggedright\arraybackslash}p{2.8cm} >{\raggedright\arraybackslash}p{2.8cm} >{\raggedright\arraybackslash}p{2.8cm}}
\toprule
\textbf{Dimension} & \textbf{SkillNet (Ours)} & \textbf{ClawHub} & \textbf{SkillsMP} & \textbf{SkillHub} & \textbf{Skills.sh} \\
\midrule
\textbf{Core Positioning} & \textbf{Full-lifecycle infrastructure} (creation, evaluation, \& connection) & npm-like version management hub & Massive open-source ecosystem catalog & Premium marketplace (curated \& stacked) & Universal skill directory \& leaderboard \\
\addlinespace
\textbf{Automated Creation} & \textbf{Supported} (generates skills from trajectories, code, and documents via LLMs) & Not supported & Not supported & Not supported & Not supported \\
\addlinespace
\textbf{Quality Evaluation} & \textbf{Multi-dimensional metrics} (safety, completeness, executability, maintainability, cost-aware) & None (relies on highlighted recommendations) & Basic filtering (e.g., GitHub stars) & Built-in LLM rating (e.g., S-rank/B-rank) & Community leaderboard (downloads/trends) \\
\addlinespace
\textbf{Relationship Analysis} & \textbf{Supported} (extracts structural relations to build a skill graph) & Not supported & Not supported & Manual pre-configured skill stacks & Not supported \\
\addlinespace
\textbf{Integration \& Distribution} & \textbf{Python SDK \& CLI tool} (\texttt{skillnet-ai}) & npx-based package manager (\texttt{clawhub install}) & GitHub repository aggregation \& CLI & Desktop client 1-click install \& CLI & Universal CLI (\texttt{npx skills add}) \\
\addlinespace
\textbf{Search Mechanism} & Vector semantic search + keyword matching & Vector semantic search + version control & AI semantic search + keyword matching & Semantic search + ranking filters & Global directory search \\
\addlinespace
\textbf{Skill Quantity} & \textbf{600k+ Total, 500k+ Curated}  & $\sim$9k+ & $\sim$261k+ & $\sim$21k+ & $\sim$71k+ \\
\bottomrule
\end{tabular}
}
\caption{Comparison of SkillNet and Existing AI Agent Skill Platforms}
\label{tab:platform_comparison}
\end{table*}

Parallel to the development of agentic systems, a number of community-driven skill repositories \cite{skillsmp2026,skillhub2026,skillssh2026,clawhub2026} and evaluations \cite{skillsbench2025,liu2026agentskillswildempirical,ling2026agentskillsdatadrivenanalysis} have emerged to curate agent skills, aiming to promote reuse, standardization, and ecosystem-level collaboration. 
For instance, platforms like ClawHub function as npm-like version management hubs, while SkillsMP and Skills.sh act as expansive open-source directories aggregating GitHub repositories. Additionally, SkillHub introduces premium marketplace features with predefined skill stacks and basic rating systems. 
Recently, \citet{li2026skillsbench} introduce SkillsBench, an 87-task, 11-domain benchmark showing that curated agent Skills significantly boost LLM agent performance (+16.2 pp on average) while self-generated Skills offer no gain, revealing that models benefit from consuming but cannot reliably author procedural knowledge.
These platforms provide valuable resources for sharing prompts, tool wrappers, and task-solving templates.

As summarized in Table \ref{tab:platform_comparison}, existing platforms predominantly operate as static package managers or marketplaces, facing three critical limitations:
First, they largely rely on manual curation and ad hoc quality control, lacking automated mechanisms to generate skills dynamically from agent trajectories or existing codebases. 
Second, their evaluation practices primarily focus on simple community metrics (e.g., repository stars) or end-task performance, failing to offer comprehensive insights into intrinsic properties such as safety, completeness, executability, maintainability, and cost-awareness. 
Third, these skill collections often suffer from redundancy, brittleness, and poor composability because skills are treated as isolated entities, restricting their scalability in large agent populations.

These limitations motivate the need for a unified framework for skill organization, verification, and evolution. 
In contrast to distribution-centric platforms, our proposed \textbf{SkillNet} provides a full-lifecycle infrastructure. 
It addresses the aforementioned bottlenecks through automated skill creation via LLM pipelines, rigorous multi-dimensional evaluation, and relational connectivity that weaves isolated tools into a structured Skill Graph, thereby offering comprehensive foundational support for self-evolving agent ecosystems.

\section{Conclusion, Discussion and Future Work}
In this work, we introduced SkillNet, an open infrastructure for creating, evaluating, and organizing AI skills at scale. By systematically consolidating experience, structuring skills, and providing principled evaluation, SkillNet enables agents to improve cumulatively, perform reliably across tasks, and adapt to complex, open-ended environments. This framework lays the foundation for scalable continual learning and robust skill composition, bridging the gap between raw model capability and sustained, evolving intelligence.

Conceptually, SkillNet is grounded in a unified view of three complementary constraints on an agent’s generative capability: workflows, memory, and skills. Workflows impose explicit procedural structure, ensuring reliability but remaining inherently rigid. Memory accumulates contextual experience and associative knowledge, enabling adaptation but lacking operational boundaries. Skills bridge these extremes by packaging reusable capability units that both constrain generation and organize memory into actionable patterns. 
In this perspective, skills serve as the structured interface through which memory becomes executable and workflows become flexible.

Looking ahead, the emerging paradigm of the ``one-person company'' or ``one-person lab'' envisions a single expert orchestrating a society of agents. 
Through SkillNet, skills become the primary unit of knowledge integration and delegation: individuals curate skill repositories, agents compose them into workflows, and memory continuously refines them through experience. 
This closed loop transforms isolated automation into cumulative machine expertise, enabling minimal human teams to achieve organizational-scale intelligence.

\paragraph{Open-World Skill Evolution.}
Achieving automatic skill discovery, abstraction, and cross-domain transfer in open-world settings remains highly challenging. 
In domains such as \textbf{industrial manufacturing, finance, and scientific research}, this requires the dynamic composition and optimization of complex tasks. 
An industry-specific, privately curated SkillNet may itself become a foundational component of the agent infrastructure.
Moreover, integrating skill evolution mechanisms with online feedback, causal reasoning, and uncertainty modeling holds promise for improving the reliability of skill selection.

\paragraph{Model-Skill Synergy.}
Although SkillNet provides agents with large-scale executable skills, the synergy between these skills and the underlying model capabilities remains largely unexplored. 
In particular, the question of how to leverage neuro-symbolic integration and memory mechanisms to enable skill structures to guide model decision paths and to dynamically restructure skill hierarchies and dependencies as model capabilities evolve remains central to systematic study.

\paragraph{Multi-Agent Collaboration and Knowledge Sharing.}
In multi-agent environments, SkillNet can function as a shared representation and exchange layer, supporting collaborative planning, knowledge transfer, and the accumulation of experience across agents.
By continuously consolidating agent behaviors into reusable skills, SkillNet may further supports the emergence of \textbf{digital avatar} whose capabilities are progressively distilled from accumulated skills.
More broadly, this paradigm may open a pathway toward collective intelligence, where skills evolve into transferable, composable units of coordination and where digital personas can inherit, share, and extend competencies beyond individual agents.

\section{Limitations}
There remain several limitations in the present work.
First, the coverage of skills is inevitably incomplete. Many capabilities in private or specialized domains cannot be incorporated, while low-frequency or highly tacit abilities are difficult to capture and consolidate within the repository, particularly when they resist explicit linguistic description.
Second, the quality of self-constructed skills cannot be fully guaranteed. 
Although our evaluation procedures filter out some problematic cases, a substantial portion of skills still lacks rigorous and systematic assessment.
If malicious users contribute ``poisoned'' or adversarial skills, SkillNet’s current Safety evaluation mechanism can detect some of these cases, but cannot fully mitigate them.
Third, an end-to-end pipeline that translates natural-language requirements into fully instantiated agents via SkillNet has not yet been established; this remains an important direction for future work.
We will continue to curate, maintain, and improve the system over time.

\section{Acknowledgement}
We sincerely thank the open-source developer community on GitHub (e.g., Skill\_Seekers\footnote{\url{https://github.com/yusufkaraaslan/Skill_Seekers}}, Vercel\_Skills\footnote{\url{https://github.com/vercel-labs/skills}}) for their broad and generous contributions to skills. 
We are grateful to Lu Chen, Xiang Li, Zhuosheng Zhang for their valuable suggestions and insightful feedback on this work.
% \section{Limitations}

\bibliographystyle{unsrtnat}  
\bibliography{skillnet} 

\clearpage

\beginappendix

\section{Comparison Traits Details}
\label{appdix:comparison}
We compare representative agentic skill benchmarks along eight traits spanning three dimensions: benchmark construction, benchmark scope, and evaluation focus.

\begin{itemize}
    \item \textbf{Real-World Source}: Whether the benchmark is grounded in artifacts, repositories, or tasks drawn from real-world or community environments, rather than being primarily hand-crafted, synthetic, or narrowly curated for evaluation convenience.
    
    \item \textbf{Automatic Construction}: Whether the benchmark can be constructed largely through an automated pipeline, such as automatic data collection, graph construction, task synthesis, or verifier generation, instead of relying mainly on manual authoring and curation.
    
    \item \textbf{Dynamic Extensibility}: Whether the benchmark can be incrementally updated as new skills, artifacts, or workflows emerge, allowing it to evolve with the underlying ecosystem rather than remaining a static snapshot.
    
    \item \textbf{Domain Coverage}: Whether the benchmark covers a sufficiently broad range of domains and task settings, enabling evaluation beyond a narrow vertical such as software engineering or a small fixed taxonomy.
    
    \item \textbf{Skill Pool}: Whether the benchmark includes a comparatively large candidate pool of skills or skill-like artifacts, such that agents must operate under realistic retrieval or selection conditions rather than relying only on pre-attached or task-specific skills.
    
    \item \textbf{Task Numbers}: The number of benchmark task instances available for evaluation. Larger task sets typically improve statistical reliability, diversity, and robustness of conclusions, although scale alone does not imply realism or breadth.
    
    \item \textbf{Skill Construction}: Whether the benchmark explicitly evaluates an agent's ability to create, generate, or refine new skills, rather than only consuming pre-existing skills during task solving.
    
    \item \textbf{Skill Composition}: Whether the benchmark explicitly evaluates how agents retrieve, organize, or compose multiple skills or tool-like units into coherent workflows, rather than treating skills as isolated single-use aids.
    
    \item \textbf{Skill Execution}: Whether the benchmark evaluates end-to-end task solving with executable environments and deterministic verification, focusing on whether skill use ultimately improves real execution outcomes.
\end{itemize}

\section{Further Experiment Details}
\subsection{Construction Baselines}
We additionally evaluate SkillSeeker \cite{skillseekers} under skill construction setting. 
The results are reported in Table~\ref{tab:construction_baseline}. The results suggest that different skill construction methods can play complementary roles across models. Meanwhile, both methods largely rely on directly compressing documentation into reusable skills, their effectiveness is inherently limited. This observation points to a promising direction for future work: skill construction methods that go beyond static distillation and instead support iterative self-improvement and empirical validation.

\subsection{End-to-End Performance in the Wild}
We evaluate agents in a wild setting, where they are exposed to a large skill library containing over 2000 skills, and measure their task performance.
The results are shown in Table~\ref{tab:in_wild}. 
Compared with the setting where the agent is provided with the correct official skills, performance drops on both Easy and Hard tasks to varying degrees, with the degradation being particularly pronounced on Hard tasks. 
We attribute this decline to the difficulty of operating over a massive skill library: the agent often fails to retrieve relevant skills, or retrieves only an incomplete subset of them. 
These results further highlight that a strong skill composition baseline is crucial for enabling effective skill usage in realistic, open-world scenarios.

\begin{table}[ht!]
\centering

\begin{minipage}{0.48\textwidth}
\centering

\begin{tabular}{@{}lcccc@{}}
% \begin{table}[ht!]
    \centering
    \newcolumntype{C}{>{\centering\arraybackslash}p{1.1cm}}
    % \resizebox{0.99\textwidth}{!}{
    % \begin{tabular}{l|CC|CC}
    \setlength{\tabcolsep}{2pt}
    \begin{tabular}{@{}lcccc@{}}

    \toprule
    \multirow{2}{*}{\textbf{Models}}
    & \multicolumn{2}{c}{\textit{Skill Creator}}
    & \multicolumn{2}{c}{\textit{Skill Seeker}}\\
    \cmidrule(lr){2-3} \cmidrule(lr){4-5}
    & \textit{Easy} & \textit{Hard}
    & \textit{Easy} & \textit{Hard}
    \\
    \midrule

    % \rowcolor{mydarkbrown!40}
    % \multicolumn{5}{c}{\textit{\textbf{Claude Code}}} \\
    \multicolumn{5}{c}{
    \cellcolor{mydarkbrown!40}
    \textit{\textbf{Claude Code}}
    }\\
    Minimax-M2.7 & 66.88 & 20.73 & 60.39 & 23.17\\
    Kimi-K2.6 & 73.38 & 31.10 & 75.32 & 38.41\\
    Deepseek-V4 & 68.83 & 29.88 & 72.73 & 23.17\\
    GLM-5.1 & 79.22 & 31.10 & 70.78 & 28.05 \\

    \bottomrule
    \end{tabular}%
    % }
\label{tab:construction_baseline}
% \end{table}
\end{tabular}
\captionof{table}{Performance of skill construction baselines.}
\label{tab:construction_baseline}

\end{minipage}
\hfill
\begin{minipage}{0.48\textwidth}
\centering

\begin{tabular}{@{}lcccc@{}}
% \begin{table}[h]
    \centering
    \newcolumntype{C}{>{\centering\arraybackslash}p{1.1cm}}
    % \resizebox{0.99\textwidth}{!}{
    % \begin{tabular}{l|CC|CC}
    \setlength{\tabcolsep}{2pt}
    \begin{tabular}{@{}lcccc@{}}

    \toprule
    \multirow{2}{*}{\textbf{Models}}
    & \multicolumn{2}{c}{\textit{Official Skill}}
    & \multicolumn{2}{c}{\textit{In Wild}}\\
    \cmidrule(lr){2-3} \cmidrule(lr){4-5}
    & \textit{Easy} & \textit{Hard}
    & \textit{Easy} & \textit{Hard}
    \\
    \midrule

    % \rowcolor{mydarkbrown!40}
    % \multicolumn{5}{c}{\textit{\textbf{Claude Code}}} \\
    \multicolumn{5}{c}{
    \cellcolor{mydarkbrown!40}
    \textit{\textbf{Claude Code}}
    }\\
    Minimax-M2.7 & 65.58 & 26.22 & 65.58 & 23.17\\
    Kimi-K2.6 & 74.03 & 34.15 & 66.23 & 25.61 \\
    Deepseek-V4 & 69.44 & 29.87 & 68.83 & 24.39\\
    GLM-5.1 & 71.43 & 32.93 & 66.23 & 24.39\\

    \bottomrule
    \end{tabular}%
    % }
\label{tab:in_wild}
% \end{table}
\end{tabular}

\captionof{table}{End-to-End Task Performance in the Wild Setting.}
\label{tab:in_wild}

\end{minipage}

\end{table}

% \begin{table}[h]
%     \centering
%     \newcolumntype{C}{>{\centering\arraybackslash}p{1.1cm}}
%     % \resizebox{0.99\textwidth}{!}{
%     % \begin{tabular}{l|CC|CC}
%     \setlength{\tabcolsep}{2pt}
%     \begin{tabular}{@{}lcccc@{}}

%     \toprule
%     \multirow{2}{*}{\textbf{Models}}
%     & \multicolumn{2}{c}{\textit{Official Skill}}
%     & \multicolumn{2}{c}{\textit{In Wild}}\\
%     \cmidrule(lr){2-3} \cmidrule(lr){4-5}
%     & \textit{Easy} & \textit{Hard}
%     & \textit{Easy} & \textit{Hard}
%     \\
%     \midrule

%     \rowcolor{mydarkbrown!40}
%     \multicolumn{5}{c}{\textit{\textbf{Claude Code}}} \\
%     Minimax-M2.7 & 65.58 & 26.22 & 65.58 & 23.17\\
%     Kimi-K2.6 & 74.03 & 34.15 & 66.23 & 25.61 \\
%     Deepseek-V4 & 69.44 & 29.87 & 68.83 & 24.39\\
%     GLM-5.1 & 71.43 & 32.93 & 66.23 & 24.39\\

%     \bottomrule
%     \end{tabular}%
%     % }
% \caption{End-to-End Task Performance of Agents in the Wild Setting.}
% \label{tab:in_wild}
% \end{table}

\section{Benchmark Details}
\label{appdix:benchmark_details}
\subsection{Benchmark Construction Details}
\paragraph{SkillNet Construction Details.}
We implement SkillNet construction as a staged pipeline that turns broad community skill retrieval results into a scenario-mediated directed skill graph. The pipeline first collects and materializes candidate skills, removes near-duplicates, abstracts each skill into pre- and post-scenarios, aligns compatible scenarios across skills, and finally aggregates verified alignments into \texttt{compose\_with} edges for task synthesis.

\textbf{Skill Retrieval and Materialization.}
SkillNet construction starts from broad query seeds covering diverse domains. For each query, we call the SkillNet search API in vector mode with a fixed similarity threshold and bounded top-$k$ limit. Retrieved skills are ranked by GitHub stars and truncated to the top candidates, with optional deduplication by skill URL or name. The selected \texttt{skill\_url}s are then resolved and downloaded from GitHub, producing a local skill corpus with a manifest that records provenance and download status.

\textbf{Near-Duplicate Skill Removal.}
To reduce redundancy, each retained skill is embedded using its name, metadata, matched queries, and full \texttt{SKILL.md} content. A FAISS nearest-neighbor index identifies highly similar skills, which are merged with union-find clustering. The representative of each cluster is selected by GitHub stars with deterministic tie-breaking.

\textbf{Scenario Extraction and Canonicalization.}
For each deduplicated skill, an LLM extracts scenario-mediated transitions from the skill specification. A pre-scenario describes the state before the skill is applicable, while a post-scenario describes the state after successful execution. Raw scenarios are normalized, embedded, and organized into a sparse FAISS similarity graph. Louvain community detection provides coarse groups, which are further refined by complete-linkage clustering. Each final cluster is assigned a canonical scenario name and linked back to its skills.

\textbf{Scenario Alignment and Edge Verification.}
Cross-skill edge candidates are induced by retrieving compatible pre-scenarios for each post-scenario. Post-scenario embeddings query a FAISS index over pre-scenario embeddings, and candidates below the retrieval threshold are discarded. An LLM judge then verifies whether the source post-scenario can naturally satisfy the target pre-scenario in a real workflow.

\textbf{Graph Assembly.}
The graph builder aggregates verified post-to-pre alignments into directed \texttt{compose\_with} edges between skill nodes. Multiple alignments between the same source--target skill pair are grouped into one edge with accumulated evidence. Each edge stores the source and target skills, supporting scenarios, alignment types, confidence scores, retrieval scores, and scenario-level evidence. The final directed SkillNet graph is used for subgraph sampling and downstream task synthesis.

\paragraph{Quality Control for SkillNet.}
\label{appdix:quality}
To ensure that SkillNet remains reliable and non-redundant, we apply quality control throughout its construction, covering skills, edges and the downstream artifacts.
% entity files and the source documents used to derive skills. 
\textit{1) Skill Level:} Each downloaded \texttt{SKILL.md} is evaluated by an LLM judge along three dimensions: environment cost, output verifiability, and documentation quality. A skill is retained only if it meets the minimum threshold for each dimension and also satisfies the overall weighted score requirement. We further remove near-duplicate skills by embedding the full skill descriptions, retrieving high-similarity neighbors with FAISS, clustering duplicates, and keeping the highest-star representative within each cluster. 
\textit{2) Edge Level:} Candidate skill pairs are reviewed using both their full \texttt{SKILL.md} files and scenario-level evidence. We remove edges when the target skill merely restates the source capability, represents an alternative implementation of the same function, or fails to introduce a substantively new state transition. 
\textit{3) Artifact Level:} Auxiliary documents and files are retained only if they are substantively relevant to the associated skill, which helps filter out loosely related or noisy context. 
Together, these steps keep SkillNet compact, well-grounded, and effective for downstream task synthesis.

\paragraph{Quality Control for Synthesized Tasks.}
Task-level quality control covers four core components 349
of each synthesized task: the instruction, the refer- 350
ence solution, the test script, and the task difficulty.

\textbf{Human-Aligned Rubric Judge for Task Instructions.}
We design a rubric-based judge to assess the quality of synthesized task instructions along three key dimensions: \textit{1) Verifiability}, whether the instruction admits a uniquely testable outcome under the evaluation script;
\textit{2) Human-expert Realism}, whether the instruction resembles one that a human expert would naturally write, rather than exhibiting an overly synthetic or machine-generated style; and \textit{3) DAG compliance}, whether the instruction is consistent with the workflow specified by the sampled DAG.
To further align the judge with authentic human preferences, we refine the prompting strategy of the LLM-as-a-Judge using manually curated benchmarks, such as SkillsBench, as supervision signals.

\textbf{PRM-Based Verification of Ground-Truth Solutions.}
Although we obtain candidate solutions by prompting an agent with meta guidance, the resulting outputs are often difficult to verify directly and may otherwise require costly review by human experts. 
To address this issue, we employ an additional agent to monitor and assess the intermediate execution trajectory used to produce the solution. 
If this verifier agent judges the reasoning and execution process to be sound, we treat the resulting artifact as a trustworthy ground-truth solution.
Importantly, we do not immediately discard samples whose trajectories are flagged as problematic, as doing so would incur substantial synthesis cost. 
Instead, we feed the failure signals and diagnostic feedback from unsuccessful attempts into the context of subsequent retries. This iterative refinement process continues until a valid solution is produced or a maximum retry budget is reached, at which point the sample is abandoned.

\textbf{Test Script Generation with Post-hoc Verification.}
The synthesized evaluation script should satisfy two key requirements: it must be fully passed by the reference solution, and it must comprehensively cover all evaluation conditions implied by the task instruction. We iteratively refine the test script until these criteria are satisfied.
In addition, to prevent the tests from being either too weak or overly restrictive, we design a rubric that allows an LLM to evaluate the quality of the test script itself. We also conduct preliminary rollouts with candidate models to assess whether the tests are under-constraining and fail to distinguish correct solutions from incomplete or incorrect ones.

\textbf{Difficulty-Aware Filtering and Human Cross-Validation.}
To ensure that the synthesized tasks genuinely challenge an agent's ability to orchestrate complex skill workflows, we adopt a rigorous two-stage task retention pipeline. 
First, we apply difficulty-aware filtering to remove tasks that can be successfully solved by GLM-5.1 \cite{glm5team2026glm5vibecodingagentic} in three independent trials without skills.
This step removes shortcut-prone tasks that can be completed from general prior knowledge alone.
The remaining tasks are then independently reviewed by three annotators with at least master's-level training in computer science. 
Each annotator evaluates instruction clarity, ground-truth correctness, and test-script reliability. Disagreements are resolved by majority vote after discussion. 
Inter-annotator agreement is high, with Fleiss's Kappa $\kappa=0.834$, indicating strong consistency within the range \(0.80 \leq \kappa \leq 1.00\).

\end{document}